\documentclass[sigconf]{acmart}

\usepackage{amsmath,amsfonts,bm}

\def\eqref#1{equation~\ref{#1}}

\def\1{\bm{1}}

\def\rvz{{\mathbf{z}}}

\def\vm{{\bm{m}}}

\def\vp{{\bm{p}}}

\def\vx{{\bm{x}}}

\def\vz{{\bm{z}}}

\DeclareMathAlphabet{\mathsfit}{\encodingdefault}{\sfdefault}{m}{sl}
\SetMathAlphabet{\mathsfit}{bold}{\encodingdefault}{\sfdefault}{bx}{n}

\usepackage{bbding}
\usepackage{hyperref}       
\usepackage{url}            
\usepackage{amsfonts}       
\usepackage{nicefrac}       
\usepackage{microtype}      
\usepackage{xcolor}         

\usepackage{float}
\usepackage{graphicx}
\usepackage{color, soul}
\usepackage{wrapfig}
\usepackage{amsmath, bbm}
\usepackage{multirow}
\usepackage{enumitem}
\usepackage[normalem]{ulem}
\useunder{\uline}{\ul}{}

\AtBeginDocument{%
  }

\setcopyright{acmcopyright}

\copyrightyear{2023}
\acmYear{2023}
\setcopyright{acmlicensed}\acmConference[KDD '23]{Proceedings of the 29th ACM SIGKDD Conference on Knowledge Discovery and Data Mining}{August 6--10, 2023}{Long Beach, CA, USA}
\acmBooktitle{Proceedings of the 29th ACM SIGKDD Conference on Knowledge Discovery and Data Mining (KDD '23), August 6--10, 2023, Long Beach, CA, USA}
\acmPrice{15.00}
\acmDOI{10.1145/3580305.3599343}
\acmISBN{979-8-4007-0103-0/23/08}

\begin{document}

\title{Feature-based Learning for Diverse and Privacy-Preserving Counterfactual Explanations}


\author{Vy Vo}
\affiliation{%
  \institution{Monash University, Australia}
  \institution{CSIRO’s Data61, Australia}
  \country{}
}
\email{Tran.Vo@monash.edu}

\author{Trung Le}
\affiliation{%
  \institution{Monash University, Australia}
  \country{}
}

\author{Van Nguyen}
\affiliation{%
  \institution{Monash University, Australia}
  \institution{CSIRO’s Data61, Australia}
  \country{}
}

\author{He Zhao}
\affiliation{%
 \institution{CSIRO’s Data61, Australia}
 \country{}
 }

 \author{Edwin V. Bonilla}
\affiliation{%
 \institution{CSIRO’s Data61, Australia}
 \institution{Australian National University, Australia}
 \country{}
 }

\author{Gholamreza Haffari}
\affiliation{%
  \institution{Monash University, Australia}
  \country{}
}

\author{Dinh Phung}
\affiliation{%
  \institution{Monash University, Australia}
  \institution{VinAI Research, Vietnam}
  \country{}
}

\renewcommand{\shortauthors}{Vo et al.}

\begin{abstract}
Interpretable machine learning seeks to understand the reasoning process of complex black-box systems that are long notorious for lack of explainability. One flourishing approach is through counterfactual explanations, which provide suggestions on what a user can do to alter an outcome. Not only must a counterfactual example counter the original prediction from the black-box classifier but it should also satisfy various constraints for practical applications. Diversity is one of the critical constraints that however remains less discussed. While diverse counterfactuals are ideal, it is computationally challenging to simultaneously address some other constraints. Furthermore, there is a growing privacy concern over the released counterfactual data. To this end, we propose a feature-based learning framework that effectively handles the counterfactual constraints and contributes itself to the limited pool of private explanation models. We demonstrate the flexibility and effectiveness of our method in generating diverse counterfactuals of actionability and plausibility. Our counterfactual engine is more efficient than counterparts of the same capacity while yielding the lowest re-identification risks. 
\end{abstract}



\maketitle
\section{Introduction}\label{intro}
The eminence of deep neural networks in recent years has proliferated the use of machine learning in various real-world applications. Such models provide remarkable predictive performance yet often at a cost of transparency and interpretability. This has sparked controversy over whether to rely on algorithmic predictions for high-stakes decision making such as graduate admission \cite{waters2014grade,acharya2019comparison}, job recruitment \cite{ajunwa2016hiring}, credit assessment \cite{lessmann2015benchmarking} or criminal justice \cite{gifford2018legal,nguyen2021information}. Progress in interpretable machine learning offers interesting solutions to explaining the underlying behavior of black-box models \citep{ribeiro2016should,nguyen2022information, vo2023an}. One useful approach is through counterfactual examples\footnote{Counterfactual explanation is sometimes referred to as algorithmic recourse.}, which sheds light on what modifications to be made to an individual's profile that can \underline{counter} an unfavorable decision outcome from a black-box classifier. Such explanations explore what-if scenarios that suggest possible recourses for future improvement. Counterfactual explainability indeed has important social implications at both personal and organizational levels. For instance, feedback like `getting $1$ more referral' or `being fluent in at least $2$ languages' would help unsuccessful candidates better prepare for future job applications. By advocating for transparency in decision making, organizations can improve their attractiveness to top talents while inspecting possible prejudice implicitly introduced in historical data and consequentially embedded in the classifiers producing biased decisions.   

The ultimate goal of this line of research is to provide realistic guidelines as to what actions an individual can take to achieve a desired outcome. Desiderata of counterfactual explanations have been extensively discussed in previous literature \citep{karimi2020survey,verma2020counterfactual,guidotti2022counterfactual,verma2022amortized}. To be of practical use, a counterfactual explanation should at least satisfy the following characteristics:

\begin{itemize}[leftmargin=*]
    \item \textit{Validity:} By definition, a counterfactual example must change the original black-box outcome to a desired one.
    \item \textit{Sparsity:} Counterfactuals should be close to the original example where a minimal number of features are modified.     
    \item \textit{Actionability:} Counterfactual explanations should only suggest actionable or feasible changes. In particular, changes should be made on mutable features e.g., Work Experience or SAT scores, while leaving immutable features unchanged e.g., Gender or Ethnicity.
    \item \textit{Diversity:} Diverse explanations are preferable to capture different preferences from the same user so that they can freely explore multiple options to select the best fit. 
    \item \textit{Plausibility}\footnote{The terms \textit{plausibility} and \textit{feasibility} are often used interchangeably.}: Plausible or realistic counterfactuals are to obey the input domain and constraints within/among features. For example, Age cannot decrease or be above $200$. 
    \item \textit{Scalability:} Inference should be done simultaneously and efficiently for multiple input examples.
\end{itemize}
 
Among these desiderata, \textit{diversity} emerges as a non-trivial property to address. Given an instance, a diverse counterfactual engine returns a set of different counterfactual profiles that should all lead to the desired outcome. Ensuring that the entire explanation set satisfies \textit{validity} while dealing with constraints given by \textit{actionability} and \textit{plausibility} poses a computational challenge. \textit{Scalability} becomes another important consideration mainly due to the fact that most of the existing approaches process counterfactuals separately for each input data point. Furthermore, strongly enforcing \textit{sparsity} results in a smaller subset of features that can be changed. This hence can compromise \textit{diversity} since we expect counterfactual states to differ from one to another substantially. On the other hand, there has been a growing concern over the privacy risks of model explanation \cite{milli2019model,sokol2019counterfactual,shokri2021privacy}. \citet{aivodji2020model} points out that diverse counterfactual explanations make the system more vulnerable as the released examples reveal the model decision boundaries and could disclose sensitive information such as health conditions or financial data. A \textit{linkage attack} is one such malicious attempt, which refers to the action of recovering the identity (i.e., re-identifying) of an anonymized record in the published dataset using background knowledge. It is often done by linking records to an external dataset of the population based on the combination of several attributes \cite{samarati1998protecting,machanavajjhala2007diversity,el2008protecting}. Netflix \$1M Machine Learning Contest is a notorious data breach, in which the company disclosed a dataset of $100$ million subscribers with their movie ratings and preferences. \citet{narayanan2006break} revealed a successful attack of $68\%$ that was easily achieved by cross-referencing the users' dates and precise ratings of $2$ movies with a non-anonymous dataset published by IMDb (Internet Movie Database). 

Despite an overwhelming number of counterfactual explanation approaches, only a few works tackle diverse counterfactual generation \cite{russell2019efficient,karimi2021algorithmic,mothilal2020explaining,bui2022counterfactual,redelmeier2021mcce}. However, the
trade-offs between \textit{diversity} and the aforementioned constraints, including privacy protection, have not been well studied in previous papers (See Table \ref{tab:comparison} for comparison). Filling this gap, our work proposes a novel learning-based framework that effectively addresses all the above desiderata while mitigating the re-identification risk. From a methodological perspective, \textbf{our method diverges markedly from existing approaches in the following ways:} 

Firstly, we reformulate the combinatorial search task into a stochastic optimization problem to be solved via gradient descent. Unlike most previous methods that perform optimization per-input basis, we employ amortized inference to generate diverse counterfactual explanations efficiently. Amortization has been previously adopted wherein a counterfactual generative distribution is modelled via Markov Decision Processes \cite{verma2022amortized} or Variational Auto-encoders \cite{mahajan2019preserving,pawelczyk2020learning,downs2020cruds}. On one hand, none of these amortized methods addresses \textit{diversity}. On the other, we here take a different approach: we construct a learnable generation module that directly models the conditional distributions of individual features such that they form a valid counterfactual distribution when combined. 

Another point of difference of ours lies in the usage of Bernoulli sampling to ensure \textit{sparsity}. In prior works, standard metrics such as $\mathrm{L1}$ or $\mathrm{L2}$ are often used to penalize the distance between the counterfactual and original data point. \citet{verma2020counterfactual} criticizes this approach as unnatural, especially for categorical features. Avoiding the use of distance measures, we optimize along a feature selection module to output the likelihood of the feature being mutated. This module can be adapted to any user-defined constraints about the mutability of features.

Finally, we go beyond existing approaches by tackling the constraint of privacy preservation exposed to diverse explanations. The key strategy is to discretize continuous features and operate the counterfactual generation engine in the categorical feature space. Discretization is closely related to the generalization technique used in privacy-preserving data mining (PPDM) \cite{el2008protecting,mendes2017privacy}. It is also treated as a subroutine to analyze the composition of differential privacy algorithms \cite{gopi2021numerical,ghazi2022faster}. The idea is that it is by nature easier to uniquely identify a profile based on continuous features, so discretization is expected to increase the quantities of profiles linked back to a certain group of attributes. Another defense effort against linkage attack can be found in \cite{goethals2022privacy}. The paper proposes an algorithm named CF-K that heuristically searches for an equivalence class for each counterfactual instance such that every record is indistinguishable from at least $k-1$ others. CF-K is viewed as an add-on that theoretically can be implemented on top of any counterfactual generative system. However, in practice, this strategy is extremely expensive since it requires repetitively querying the model explainer for a possibly larger number of counterfactuals than requested. Though sharing the same motivation, we here contribute a counterfactual explanation model with a built-in privacy preservation functionality. 

Our contributions can be summarized as follows:
\begin{itemize}[leftmargin=*]
\item We introduce \textbf{L}earning to \textbf{C}ounter (\textbf{L2C}) - a stochastic feature-based approach for learning counterfactual explanations that address the counterfactual desirable properties in a single end-to-end differentiable framework. 
\item Through extensive experiments on real-world datasets, L2C is shown to balance the counterfactual trade-offs more effectively than the existing methods and achieve diverse explanations with the lowest re-identifiability risk. To the best of our knowledge, L2C is the first amortized engine that supports diverse counterfactual generations with privacy-preservation capability. 
\end{itemize}  
\raggedbottom

\begin{table*}[!h]
  \caption{Desiderata comparison of related counterfactual explanation methods. 
  $^*$Privacy refers to whether the output data is protected against linkage attack or re-identification risk. L2C satisfies all of these critical constraints.}
  \vspace{-0.5em}
  \label{tab:comparison}
  \resizebox{0.6\textwidth}{!}{
  \begin{tabular}{l | c c c c c c}

    \toprule
    Method & Sparsity & Actionability & Diversity & Plausibility & Scalability & Privacy$^*$ \\
    \midrule
    \textbf{L2C (Ours)} & \Checkmark & \Checkmark & \Checkmark & \Checkmark & \Checkmark & \Checkmark \\
    DICE \cite{mothilal2020explaining} & \Checkmark & \Checkmark & \Checkmark &  &  &  \\
    COPA \cite{bui2022counterfactual} & \Checkmark & \Checkmark & \Checkmark &  &  & \\
    MCCE \cite{redelmeier2021mcce} & \Checkmark & \Checkmark & \Checkmark &  & \Checkmark &  \\
    Coherent CF \cite{russell2019efficient} & \Checkmark  &  & \Checkmark &  &  \\
    MACE \cite{karimi2020model} & \Checkmark & \Checkmark & \Checkmark &  &  &  \\
    MOC \cite{dandl2020multi} & \Checkmark & \Checkmark &  & \Checkmark  & &  \\
    CERTIFAI \cite{sharma2020certifai} &  &  \Checkmark &  &  &  &  \\
    
    Feasible-VAE \cite{mahajan2019preserving} & \Checkmark & \Checkmark &  & \Checkmark & \Checkmark &  \\
    FastAR \cite{verma2022amortized} & \Checkmark & \Checkmark &  & \Checkmark & \Checkmark &  \\
    CRUDS \cite{downs2020cruds} &  & \Checkmark &  & \Checkmark  &  \Checkmark & \\
    C-CHVAE \cite{pawelczyk2020learning} &  & \Checkmark &  &  & \Checkmark  &  \\
    CF-K \cite{goethals2022privacy} &  & &  &  & &  \Checkmark \\

    \bottomrule 
    
  \end{tabular}
  }
  
\end{table*}

\section{Related Works}\label{litreview}
Recent years have seen an explosion in the literature on counterfactual explainability, from works that initially focused on one or two characteristics or families of models to those that can deal with multiple constraints and various model types. There have been many attempts to summarize major themes of research and discuss open challenges in great depth. We therefore refer readers to \cite{karimi2020survey,verma2020counterfactual,guidotti2022counterfactual} for excellent surveys of methods in this area. We here focus on reviewing algorithms that can support diverse (or at least multiple) local counterfactual generations.

Dealing with the combinatorial nature of the task, earlier works commonly adopt mixed integer programming \cite{russell2019efficient}, genetic algorithms \cite{sharma2020certifai}, or SMT solvers \cite{karimi2020model}. Another recent popular approach is gradient-based optimization \cite{mothilal2020explaining,bui2022counterfactual}, which involves iteratively perturbing the input data point according to an objective function that incorporates desired constraints. The whole idea of \textit{diversity} is to explore different combinations of features and feature values that can counter the original prediction while accommodating various user needs. To support \textit{diversity}, \citet{russell2019efficient} in particular enforces hard constraints on the current generations to be different from the previous ones. Such a constraint will however be removed whenever the solver cannot be satisfied. Meanwhile, \citet{mothilal2020explaining} and \citet{bui2022counterfactual} add another loss term for \textit{diversity} using Determinantal Point Processes \cite{kulesza2012determinantal}, whereas the other works only demonstrate the capacity to generate multiple counterfactuals via empirical results. All of the aforementioned algorithms are computationally expensive in that input data points are handled singly and individual runs are additionally required to produce several counterfactuals. Reducing computational costs, \citet{redelmeier2021mcce} attempts to model the conditional likelihood of mutable features given the immutable features using the training data. They then adopt Monte Carlo sampling to generate counterfactuals from this distribution and filter out samples that do not meet counterfactual constraints. Given such a generative distribution, sampling of counterfactuals can therefore be done straightforwardly. Amortized optimization is another strategy to improve inference speed \cite{mahajan2019preserving,pawelczyk2020learning,downs2020cruds,verma2022amortized}.

In response to the privacy warning about model explanations \cite{milli2019model,sokol2019counterfactual,shokri2021privacy}, several defense strategies have been introduced to alleviate the risks. With strong theoretical guarantees, differential privacy \cite{dwork2006calibrating} stands out as the promising solution to preventing \textit{member inference attack} and \textit{model stealing attack} \cite{shokri2017membership, milli2019model, aivodji2020model}. With regards to linkage attacks, CF-K \cite{goethals2022privacy} is the only work we are aware of that tackles linkage attack in counterfactual explanations.


\section{Stochastic Feature-Based Counterfactual Learning}
\subsection{Problem setup}
Let $\mathcal{X}$ denote the input space where $\vx = [x_i]_{i=1}^N$ is an input vector with $N$ features of both continuous and categorical types. As discussed previously, we discretize the continuous features into equal-sized buckets, which gives us an input of $N$ categorical features wherein each feature $x_i$ has $c_i$ levels. We apply one-hot encoding on each feature and flatten them into a single input vector $\vz \in \{0,1\}^D$ where $D = \sum^{N}_{i=1} c_i$. Concretely, feature $x_i$ is now represented by the vector $\vz_i \in \mathbb{O}_{c_i}$ where the set of one-hot vectors $\mathbb{O}_{c_i}$ is defined as $\{0, 1\}^{c_i} : \sum_{j=1}^{c_i} z_{ij}= 1$. 

Let $f$ be the black-box classifying function and $y = f(\vx)$ be the decision outcome on the input $\vx$. A valid counterfactual example $\widetilde{\vx}$ associated with $\vx$ is one that alters the original outcome $y$ into a desired outcome $y' \ne y$ with  $y' = f(\widetilde{\vx})$. Let $\widetilde{\vz}$ denote the corresponding one-hot representation of $\widetilde{\vx}$. 

Actionability indicates that some features can be \textit{mutable} (i.e., changeable), while others should be kept \textit{immutable} (i.e., unchangeable). Without loss of generality, let us impose an ordering on the set of $N$ features such that the first $K$ features are mutable features (i.e., the ones that can be modified) and denote $\mathbb{K} := \{1, ..., K \} \subset \{1, ..., N\}$. For each mutable feature (i.e., $x_i$ or the one-hot vector $\vz_i$ with $i\in \mathbb{K}$), we aim to learn a local feature-based perturbation distribution $P(\widetilde{\vz}_i \mid \vz)$ where $\widetilde{\vz_i} \in \mathbb{O}_{c_i}$, while leaving the immutable features unchanged.  

It is worth noting that our method functions equally well on heterogeneous data where only categorical features are one-hot encoded while continuous features are retained at their original values. However, we believe that performing data discretization (or generalization in terms of PPDM) initially and deploying the classifiers in the discrete feature space would provide better privacy protection. For the purpose of comparing our prototype with existing approaches, we follow the standard practice of explaining classification models trained on the mixed dataset of continuous and categorical features. To make it compatible with the discretization subroutine of our framework, we represent the prediction on a transformed (fully categorical) input vector with the prediction on the input where the categorical values associated with mutable continuous features are substituted with the middle point of the corresponding intervals. We refer to this mechanism as \textbf{one-hot decoding}, which will be detailed shortly. 

\subsection{Methodology}
\begin{figure*}[!t]
\centering
\includegraphics[width=0.63\textwidth]{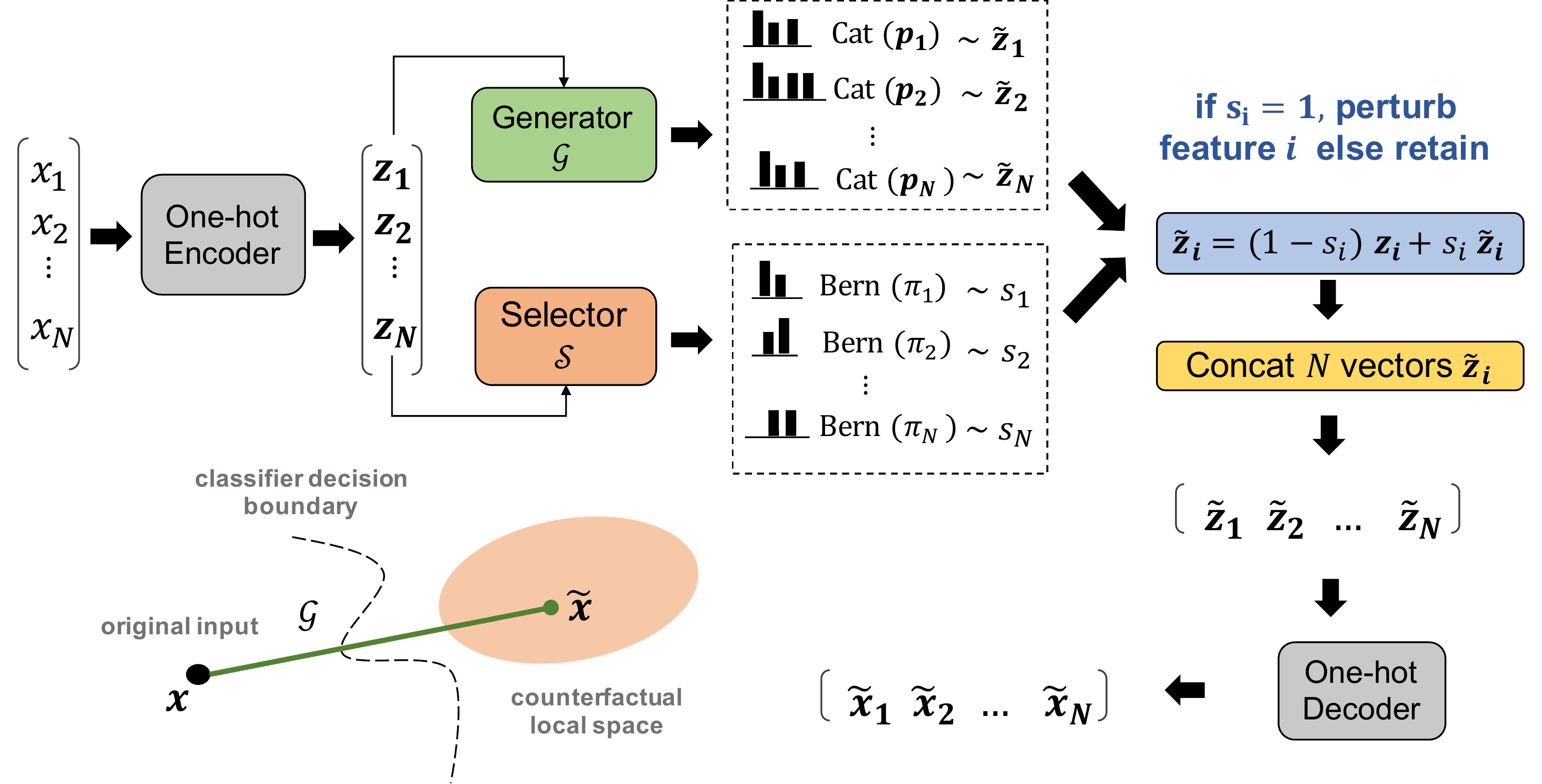}
\caption{For illustration purposes only, all features are assumed mutable in the figure. We first discretize the continuous features of input $\vx$ and one-hot encode all features into representations $\vz$. For every feature $i$, the generator $\mathcal{G}$ learns a local perturbation distribution $\textrm{Cat}(\vp_i \vert \vz )$ so that together they form a distribution of diverse counterfactual representations $\widetilde{\vz}$. Simultaneously, the selector $\mathcal{S}$ learns to output the distribution $\textrm{Multi-Bernoulli}(\boldsymbol{\pi}\vert\vz)$ capturing the probability of each feature $i$ being modified. Every feature sample pair $(\widetilde{\vz}_i, s_i)$ is passed through an operation in the blue box, which decides whether to accept the change being made to the feature $i$ given by $\widetilde{\vz}_i$. The output is then decoded into the representations $\widetilde{\vx}$ compatible with the black-box system. $\mathcal{G}$ and $\mathcal{S}$ are jointly trained via back-propagation according to Eq. (\ref{eq:new_obj}). Intuitively, $\mathcal{G}$ aims to construct a ``bridge'' across the decision boundary travelling from the input to a local space of counterfactuals.}
\label{fig:pipeline}
\vspace{-3mm}
\end{figure*}

We now detail how L2C works and addresses each counterfactual constraint. The explanation pipeline of L2C is depicted in Figure \ref{fig:pipeline}.

For each mutable feature $\vz_i$ with $i \in \mathbb{K}$, we learn a \textbf{local feature-based perturbation distribution} $P(\widetilde{\vz_i} \mid \vz)$ (i.e., $\tilde{z_i} \in \mathbb{O}_{c_i}$), which is a categorical distribution $\textrm{Cat}(\vp_i \mid \vz)$ with category probability $ \vp_i = \big[ p_{i1}, p_{i2}, ..., p_{ic_i} \big]$. We form a counterfactual example $\widetilde{\vz}$ by concatenating $\widetilde{\vz}_i \sim \textrm{Cat}(\vp_i \mid \vz)$ for the mutable features and $\vz_i$ for the immutable features. To achieve \textit{validity}, we learn the local feature-based perturbation distribution by maximizing the chance that the counterfactual examples $\widetilde{\vz}$ counter the original outcome on $\vx$. Additionally, learning local feature-based perturbation distributions over the mutable features allows us to conduct a global counterfactual distribution $P(\widetilde{\vz} \mid  \vz)$ over the counterfactual examples $\widetilde{\vz}$ defined above. Sampling from this distribution naturally leads to multiple counterfactual generations efficiently, and we also expect that individual samples $\widetilde{\vz}_i$ together can form diverse combinations of features, thereby promoting \textit{diversity} within the generative examples. 

As previously discussed, too much of \textit{diversity} can compromise \textit{sparsity}. Dealing with this constraint, for each mutable feature $\vz_i$, we propose to learn a \textbf{local feature-based selection distribution} that generates a random binary variable $s_i \sim \textrm{Bernoulli}(\pi_i\mid{\vz})$ wherein we replace $\vz_i$ by $\widetilde{\vz}_i \sim \textrm{Cat}(\vp_i \mid \vz)$ if $s_i = 1$ and leave $\tilde{\vz}_i = \vz_i$ if $s_i = 0$. Therefore, the formula to update $\widetilde{\vz}_i$ is 
$$\widetilde{\vz}_i = (1-s_i)\vz_i + s_i \widetilde{\vz}_i.$$

The benefit of having $\boldsymbol{\pi} = [\pi_i]_{i\in \mathbb{K}}$ is thus to control \textit{sparsity} by adding one more channel to decide if we should modify a mutable feature $\vz_i$. Appendix \ref{ablation:selector} presents an ablation study showing that without the selection distribution, the perturbation distribution alone can generate diverse counterfactuals yet requires changing plenty of mutable features. Meanwhile, optimizing the selection distribution jointly helps harmonize the trade-off between \textit{diversity} and \textit{sparsity}.     

\subsection{Optimization Objective}\label{optimization}
In this section, we explain how to design the building blocks of our framework L2C. As shown in Figure \ref{fig:pipeline}, our framework consists of two modules: a \textbf{counterfactual generator} $\mathcal{G}$ and a \textbf{feature selector} $\mathcal{S}$. The \textbf{counterfactual generator} $\mathcal{G}$ is used to model the feature-based perturbation distribution, while \textbf{feature selector} $\mathcal{S}$ is employed to model the feature-based selection distribution.

Specifically, given a one-hot vector representation $\vz$ of  a data example $\vx$, we feed $\vz$ to $\mathcal{G}$ to form $\mathcal{G}(\vz) = [\mathcal{G}_i(\vz)]_{i \in \mathbb{K}}$. We then apply the softmax activation function to $\mathcal{G}_i(\vz)$ to define the feature-based local distribution (i.e., $\textrm{Cat}(\vp_i \mid \vz)$)) for $\vz_i$ as

\begin{equation}\label{eq:local_dist}
    p_{ij}(\vz) = \frac{ \exp \big\{ \mathcal{G}_{ij}(\vz) \big\}}  {\sum^{c_i}_{k=1} \exp \big\{ \mathcal{G}_{ik}(\vz) \big\} }, \forall j =1,...,c_i.
\end{equation}

The module $\mathcal{S}$ takes $\vz$ to form  $\mathcal{S}(\vz) = [\mathcal{S}_i(\vz)]_{i \in \mathbb{K}}$. We then apply the Sigmoid function to $\mathcal{S}_i(\vz)$ to define the feature-based selection distribution (i.e., $\textrm{Bernoulli}(\pi_i\mid{\vz})$) for $\vz_i$ as
$$\pi_i(\vz) = \frac{1}{1 + \exp{\big\{-\mathcal{S}_i(\vz)}\big\}}.$$

To encourage \textit{sparsity} by reducing the number of mutable features chosen to be modified, we regularize $\mathcal{S}$ through $\mathrm{L1}$-norm $\Vert \boldsymbol{\pi}(\vz) \Vert_1$ with $\boldsymbol{\pi}(\vz) = [\pi_i(\vz)]_{i \in \mathbb{K}}$.

To summarize, given a one-hot vector representation $\vz$ of a data example $\vx$, we use $\mathcal{G}$ to work out the local feature-based perturbation distribution $\textrm{Cat}(\vp_i(\vz))$ for every $i \in \mathbb{K}$. We then sample $\widetilde{\vz}_i \sim \textrm{Cat}(\vp_i(\vz))$ for every $i \in \mathbb{K}$. Subsequently, we use $\mathcal{S}$ to work out the local feature-based selection distribution $\textrm{Bernoulli}(\pi_i(\vz))$ for every $i \in \mathbb{K}$. We then sample $s_i \sim \textrm{Bernoulli}(\pi_i\mid{\vz})$ and update $\widetilde{\vz}_i = (1-s_i)\vz_i + s_i \widetilde{\vz}_i$ for every $i \in \mathbb{K}$. Finally, we concatenate $\widetilde{\vz}_i$ for $i \in \mathbb{K}$ and $\vz_i$ for $i \notin \mathbb{K}$ to form the counterfactual example $\widetilde{\vz}$.  

$\mathcal{G}$ and $\mathcal{S}$ are parameterized with neural networks over total parameters $\theta$. For $\widetilde{\vz}$ to be a \textit{valid} and \textit{sparse} counterfactual associated with a desired outcome $y'$, we propose the following criterion 
\begin{equation}\label{eq:old_obj}
    \textrm{min}_{\theta} \Big[
    \mathbb{E}_{\widetilde{\vz}}\big[\textrm{CE}(f(\widetilde{\vz}), y')\big]  + \alpha \
    \mathbb{E}_{\vz} \big[ \Vert \boldsymbol{\pi}(\vz) \Vert_1 \big]
    \Big],
\end{equation}
where $f$ is the black-box function, $\textrm{CE}$ is the cross-entropy loss, $\Vert \cdot \Vert_1$ is $\mathrm{L1}$-norm, $\alpha$ is a loss weight.

\paragraph{\textbf{One-hot decoding}} Recall that $\widetilde{\vz}$ formed by concatenating many one-hot vectors is an incompatible representation to the classifier $f$, which in fact requires both continuous and one-hot features. We make a design choice of reconstructing the continuous features by taking the middle point of the range corresponding to the selected level. Specifically, the input to the one-hot decoder is $\widetilde{\vz} = [\widetilde{\vz}_i]_{i=1}^N$. If the feature $i$ originally is a categorical feature, we set $\widetilde{x}_i = \widetilde{\vz}_i$. Otherwise, we set $\widetilde{x}_i = a_i + \frac{(2k-1)(b_i - a_i)}{2c_i}$, which is the middle point of the $k$-th interval $[a_i + (k-1)(b_i - a_i)/c_i, a_i + k(b_i - a_i)/c_i]$ where $[a_i, b_i]$ is the original value range of the feature $i$ and $\widetilde{\vz}_i$ corresponds to the level $k \in \{1,...,c_i\}$ (i.e., $\widetilde{\vz}_{ik}=1$ and $\widetilde{\vz}_{ij}=0$ if $j \neq k$). Note that one-hot decoding is only applied when a continuous feature is indicated by $\mathcal{S}$ to be mutable ($s_i = 1)$. Otherwise, we revert to the original continuous value. Formally, we rewrite
\begin{equation}
 \widetilde{x}_i = (1-s_i) x_i + s_i \sum_{j=1}^{c_i}\widetilde{\vz}_{ij}\big[a_i + \frac{(2j-1)(b_i - a_i)}{2c_i}\big].   \label{eq:num_back}
\end{equation}
\raggedbottom

To assure model differentiability for training, the one-hot vector $\widetilde{\vz}_{ij}$ is relaxed into its continuous representation by using Gumbel-Softmax reparametrization trick, which is detailed in Section \ref{gumbel}. 

\textbf{The final optimization objective is now given as}
\begin{equation}\label{eq:new_obj}
    \textrm{min}_{\theta} \Big[
    \mathbb{E}_{\widetilde{\vx}}\big[\textrm{CE}(f(\widetilde{\vx}), y')\big]  + \alpha \
    \mathbb{E}_{\vz} \big[ \Vert \boldsymbol{\pi}(\vz) \Vert_1 \big]
    \Big].
\end{equation}
\raggedbottom

\paragraph{\textbf{Plausibility}} A counterfactual generative engine needs to ensure explanations are realistic for real-world applications. There are two common types of plausibility constraints: \textbf{Unary} and \textbf{Binary} monotonicity constraints. The former deals with individual features (e.g., Age cannot decrease) while the latter is concerned with the correlation of a pair of features (e.g., increasing in Education level increases Age). To handle binary constraints on heterogeneous data is not as straightforward as unary constraints. We thus delay the discussion on binary constraints until Section \ref{binary_constraint} and focus on unary constraints in the main analysis. 

Dealing with such a constraint, one can simply eliminate any counterfactuals violating the constraints during inference time. This however creates additional computational overhead and may compromise some desiderata. There are only few attempts addressing feature constraints in counterfactuals, notably \citet{mahajan2019preserving,verma2022amortized}: \citet{verma2020counterfactual} incorporate hard conditions in a Markov Decision process where a feature gets updated only if the corresponding action does not violate the constraints. Meanwhile, \citet{mahajan2019preserving} includes a hinge loss into the loss function for unary features, while specifically learning a separate linear model for every feature pair subject to a binary constraint. For L2C, the learnable local distributions can be used for this purpose conveniently. Our proposed strategy is to impose rule-based unary constraints on related features in the optimization process. Technically, for every feature to be perturbed with a non-decreasing (non-increasing) constraint, we penalize the probabilities corresponding to lower (higher) levels towards zero by multiplying them with a positive infinitesimal quantity. Concretely, given a mutable feature $i$ under monotonic constraints, let $l \in \{1, ... ,c_i\} $ denote the current state - the level corresponding to $\vz_{i}$ (i.e., $z_{il}=1$ and $z_{ij}=0$ if $j \ne l $). Let us denote the restricted set of levels as $\mathbf{C}_{i} = \{1, ..., l-1\}$ if the feature is non-decreasing and $\mathbf{C}_{i} = \{l+1, ..., c_i \}$ if it is non-increasing. The perturbation distribution $\textrm{Cat}(\vp_i \mid \vz)$ for $\vz_i$ given in Eq. (\ref{eq:local_dist}) now becomes

\begin{equation}\label{eq:constraint}
p_{ij}(\vz) \propto \varepsilon^{\boldsymbol{1}_{\mathbf{C}_i}(j)} \times \frac{ \exp \big\{ \mathcal{G}_{ij}(\vz) \big\}}  {\sum^{c_i}_{k=1} \exp \big\{ \mathcal{G}_{ik}(\vz) \big\} }, \forall j = 1,...,c_i,
\end{equation}

where $\boldsymbol{1}_{\mathbf{C}_i}(.)$ is the indicator function such that $\boldsymbol{1}_{\mathbf{C}_i}(j) = 1$ if $j \in \mathbf{C}_i $ and $\boldsymbol{1}_{\mathbf{C}_i}(j) = 0$ otherwise, meaning that the probabilities at the other levels are untouched. We here explicitly force the model to generate more samples at the higher (lower) levels while maintaining differentiability of the objective function. We choose $\varepsilon = e^{-10}$ in our experiments, but any positive value arbitrarily close to zero would suffice. 

\subsection{Reparameterization for Continuous Optimization}\label{gumbel}
Our L2C involves multiple sampling rounds back and forth to optimize the networks. To make the process continuous and differentiable for training, we adopt the reparameterization tricks \cite{jang2016categorical, maddison2016concrete}: 

\paragraph{1) Sampling $\widetilde{\vz}_i \sim \textrm{Cat}(\vp_i \mid \vz)$}:
To obtain differentiable counterfactual samples, we adopt the classic temperature-dependent Gumbel-Softmax trick \citep{jang2016categorical,maddison2016concrete}. Given the categorical variable $\rvz_i$ with category probability $\big[ p_{i1}, p_{i2}, ..., p_{ic_i} \big]$. The relaxed representation is sampled from the Categorical Concrete distribution as $\widetilde{\vz}_i \sim \textrm{Cat-Concrete}(\log p_{i1}, ..., \log p_{ic_i})$ by
$$\widetilde{z}_{ij} = \frac{\exp \big\{ (\log p_{ij}(\vz) + G_j)/\tau \big\} }{\sum^{c_i}_{k=1} \exp 
\big\{ (\log p_{ik}(\vz) + G_k)/\tau \big\}  }.$$
with temperature $\tau$, random noises $G_j$ independently drawn from Gumbel distribution $G_t =-\log(- \log u_t), \ u_t \sim \textrm{Uniform}(0,1)$. As discussed, we apply this mechanism consistently to the one-hot representations of all features. The continuous relaxation of Eq. (\ref{eq:num_back}) can be gained by simply using the one-hot relaxation $\widetilde{\vz}_i$.

\paragraph{2) Sampling $s_i \sim \textrm{Bernoulli}(\pi_i \mid {\vz})$}: 
We again apply the Gumbel-Softmax trick to relax Bernoulli variables of $2$ categories. With temperature $\tau$, random noises $G_{i0}$ and $G_{i1} \sim G_t =-\log(- \log u_t), \ u_t \sim \textrm{Uniform}(0,1)$, the continuous representation $s_i$ is sampled from Binary Concrete distribution as  $s_i \sim \textrm{Bin-Concrete}(\pi_i, 1-\pi_i)$ by
$$ s_i = \frac{\exp \{ \big( \log \pi_i(\vz) + G_{i1} \big) /\tau\}}{\exp \{ \big( \log (1-\pi_i(\vz)) + G_{i0} \big) \}/\tau \} + \exp \{ \big( \log \pi_i(\vz) + G_{i1} \big)/\tau )\} }.$$

\section{Experimental Setup}
We experiment with $4$ popular real-word datasets: German Credit \cite{Dua:2019}, Adult Income \cite{kohavi1996scaling} Graduate Admission \cite{acharya2019comparison} and Student Performance \cite{cortez2008using}. For each dataset, we select a fixed subset of immutable features based on our domain knowledge and suggestions from \citep{verma2022amortized}. We reserve the privacy analysis for German Credit and Adult Income datasets, which contain personal financial information and various attributes through which data subjects can be re-identified \cite{goethals2022privacy}. While implementing the black-box classifiers and the baseline methods, we standardize numerical features to unit variance and one-hot encode categorical features. Note again that, for our method only, we discretize numerical features into equal-sized buckets and decode the numerical features back to their original representations whenever necessary to consult the black-box model. Appendix \ref{A} describes our tasks and model design in greater detail. Our code repository can be accessed at \textbf{\url{https://github.com/isVy08/L2C/}}.  

\begin{table}[!h] 
\caption{Description of quantitative evaluation metrics. $\mathbb{C}$ denotes a set of counterfactual examples generated by an algorithmic recourse approach for a given input instance.}
\vspace{-0.5em}
\label{tab:metrics}
\resizebox{\linewidth}{!}{%
\begin{tabular}{p{2.5cm} | p{2cm}  p{6cm}}
    \toprule
    \textbf{Desiderata} & \textbf{Metric} & \textbf{Description}  \\ 
    \midrule
    \multirow{2}{*}{Validity} & \texttt{Validity} &  Proportion of samples in $\mathbb{C}$ can counter the original black-box decision outcome.  \\
    & \texttt{Coverage} & \texttt{Coverage} $= 100\%$ if there exists at least $1$ valid counterfactual in $\mathbb{C}$.  \\
    \midrule
    Sparsity\slash Actionability & \texttt{Sparsity} & Proportion of features kept unchanged, averaged over the number of samples in $\mathbb{C}$. \\
    \midrule 
    Diversity & \texttt{Diversity} & Hamming distance of a pair of counterfactual samples across all features where numerical features are discretized. The metric is averaged over all pairs of samples in $\mathbb{C}$.  \\
    \midrule
    Sparsity -- Diversity Balance  & \texttt{Harmonic mean} & F-measure of \texttt{Diversity} and \texttt{Sparsity} = $2 \cdot \texttt{Diversity} \cdot \texttt{Sparsity} \slash (\texttt{Diversity} +  \texttt{Sparsity})$. \\
    \midrule
    Plausibility & \texttt{Unary} & Proportion of examples $\mathbb{C}$ meeting the unary monotonic constraints, averaged over the number of features subject to constraints. \\
    
    \bottomrule
    \end{tabular}
}
\end{table}

\paragraph{\textbf{Performance metrics}}
Following the past works \citet{mothilal2020explaining,redelmeier2021mcce,verma2022amortized}, Table \ref{tab:metrics} outlines the commonly used metrics for quantitatively assessing the desirability of counterfactual explanations. As for \textit{diversity}, a widely adopted measure is the pairwise distance between counterfactual examples, with distance defined separately for numerical and categorical features \citep{mothilal2020explaining,redelmeier2021mcce}. Though this approach is meaningful for interpreting categorical features, we however find it quite obscure for numerical features. This motivates us to discretize numerical features again when computing \texttt{Diversity}, which captures how often a feature gets altered as well as how much the change is - specifically via how often it switches to a different categorical level. The computation of \texttt{Diversity} only considers valid counterfactuals, so if valid counterfactuals are none, \texttt{Diversity} is set to zero. It is worth noting that there fundamentally exists a trade-off between \textit{sparsity} and \textit{diversity}. To quantify how well a method can balance these two properties, we suggest taking \texttt{Harmonic mean} of \texttt{Diversity} and \texttt{Sparsity}, motivated by the development of F1-score in measuring Precision against Recall. For metrics used in privacy analysis, refer to Section \ref{result:privacy}.

\paragraph{\textbf{Evaluation setup}} We consider a general setting of binary classification where a counterfactual outcome $y'$ is opposite to the original outcome $y$, whether $y$ has label $1$ or $0$. From each method, we generate a set of $100$ counterfactual explanations. During generation, most methods, including ours, require multiple iterations of searching for the optimal set of counterfactuals based on the optimization constraints. To assure a fair comparison on efficiency, a global maximum time budget of $5$ minutes is imposed to search for a set of $100$ counterfactuals per input sample.  We compare our method top-performing baselines that support diverse counterfactual explanations: DICE~\citep{mothilal2020explaining}, MCCE \citep{redelmeier2021mcce} and COPA \citep{bui2022counterfactual}. DICE offers several search strategies: Random, KD tree, or Genetic algorithm. DICE-KDTree was consistently reported to fail across datasets \citep{verma2022amortized}, so we exclude it from our evaluation. We do not consider MACE \citep{karimi2020model} since it is extremely expensive on large datasets \citep{verma2022amortized} and often fails to converge in our experiments, nor MOC \citep{dandl2020multi} due to the lack of Python implementation. 

\section{Results and Discussion}

\subsection{Counterfactual Explanation Desiderata}\label{result:quality}
We first study whether an algorithmic recourse approach generates a set of diverse counterfactuals without sacrificing the other desiderata. Note that COPA has only been shown to work effectively on linear classifiers. Table \ref{tab:result} reports the average results over $5$ model initializations. Appendix \ref{quali} provides several illustrative examples for qualitative assessment.

Under the same time budget, our method L2C succeeds in generating $100\%$ valid counterfactuals with full coverage. Together with DICE, L2C first satisfies the most important criterion of a counterfactual explanation and resolves the trade-off against \textit{validity}. Recall that we have specified a fixed set of immutable features for each dataset, based on which we can work out the minimum sparsity threshold a counterfactual explanation should adhere to (i.e., \% immutable features). \textit{Actionability} can then be assessed by comparing \texttt{Sparsity} with this level to determine if a method satisfies the mutability of features. An adequate explanation must achieve at least this level of sparsity. MCCE evidently fails to fulfill this constraint on Adult Income and Student Performance datasets. 

Our reported results here are obtained under no other conditions than the constraints related to feature immutability and monotonicity described in Table \ref{tab:datastats}. Nevertheless, we would like to highlight the flexibility of our framework in controlling the quality of counterfactual generations during inference. Users can freely specify any sparsity threshold or additional conditions of interest to filer out unsatisfactory examples without re-training or re-optimization as in methods like DICE. Specifically, DICE employs gradient search directly on each query according to a selected set of weighting hyperparameters for each term in the objective function. To get a less sparse or more diverse example than the current generation, one needs to activate a new search routine.  

Too many constraints or too much sparsity clearly affects the diversity level of the counterfactual set. Maintaining a high \texttt{Harmonic mean} scores while satisfying almost all feature constraints demonstrates that L2C can effectively manage these trade-offs. The fact that L2C converges to valid counterfactuals with minimal violation in such a short inference time can be attributed to the practice of injecting hard constraints during optimization and global training does enhance the effect. It certainly helps circumvent the burden of heuristically eradicating violated samples. Notice also that our quantitative results align with the descriptions in Table \ref{tab:comparison}. None of the diverse counterfactual explanation approaches address \textit{plausibility} thoroughly whereas those reported to support the feasibility of features do not guarantee \textit{diversity}. Appendix \ref{ablation:amortized} provides empirical evidence for this claim, in which we compare L2C with popular amortized algorithmic recourse approaches and demonstrate our consistent superiority in generating diverse explanations efficiently without violating the required constraints (See Table \ref{tab:amortized}).

\begin{table*}[!ht]
\caption{Desirability of counterfactual explanation methods. $\downarrow$ Lower is better. $\uparrow$ Higher is better. Bold / Underline indicates the best / second-best performance for each dataset. \texttt{Time} records total inference time in seconds.}
\vspace{-0.5em}
\label{tab:result}
\resizebox{\linewidth}{!}{
\begin{tabular}{l|c c c c c c r }
\toprule
Method  &  \texttt{Sparsity (\%)}$\uparrow$ & \texttt{Diversity (\%)}$\uparrow$ &  \texttt{Harmonic Mean (\%)}$\uparrow$ & \texttt{Validity (\%)}$\uparrow$ & \texttt{Coverage (\%)}$\uparrow$ & \texttt{Unary} (\%)$\uparrow$ & \texttt{Time(s)}$\downarrow$ \\
\midrule
\multicolumn{8}{c}{German Credit (Logistic Regression) - Min Sparsity: $20.00 \%$} \\
\midrule
\textbf{L2C (Ours)} & {\ul 61.35}          & {\ul 37.31}          & \textbf{46.39}       & \textbf{100.00}      & \textbf{100.00}      & \textbf{99.06}       & {\ul 18}    \\
DICE-Random         & \textbf{88.23}       & 15.29                & 26.06                & \textbf{100.00}      & \textbf{100.00}      & {\ul 90.81}          & 1,150       \\
DICE-Genetic        & 43.45                & \textbf{37.56}       & {\ul 40.29}          & {\ul 62.87}          & {\ul 90.24}          & 56.66                & 17,615      \\
COPA                & 57.88                & 18.88                & 28.47                & 44.00                & 44.00                & 84.31                & 17,583      \\
MCCE                & 28.76                & 33.40                & 30.91                & 48.74                & \textbf{100.00}      & 58.76                & \textbf{2}  \\   

\midrule
\multicolumn{8}{c}{Adult Income (Neural Network) - Min Sparsity: $30.77 \%$} \\
\midrule

\textbf{L2C (Ours)} & {\ul 45.70}          & \textbf{28.11}       & \textbf{34.80}       & \textbf{100.00}      & \textbf{100.00}      & \textbf{97.62}       & {\ul 444}   \\
DICE-Random         & \textbf{89.26}       & 9.05                 & 16.44                & \textbf{100.00}      & \textbf{100.00}      & {\ul 87.15}          & 12,332      \\
DICE-Genetic        & 41.48                & {\ul 26.27}          & {\ul 32.14}          & {\ul 92.64}          & \textbf{100.00}      & 72.70                & 505,174     \\
MCCE                & 24.93                & 4.58                 & 7.74                 & 30.63                & 74.76                & 45.79                & \textbf{98} \\

\midrule
\multicolumn{8}{c}{Graduate Admission (Neural Network) - Min Sparsity: $14.29 \%$} \\
\midrule

\textbf{L2C (Ours)} & {\ul 42.23}          & {\ul 37.90}          & {\ul 39.94}          & \textbf{100.00}      & \textbf{100.00}      & \textbf{100.00}      & {\ul 4}     \\
DICE-Random         & \textbf{66.25}       & 30.93                & \textbf{42.15}       & \textbf{100.00}      & \textbf{100.00}      & {\ul 85.30}          & 412         \\
DICE-Genetic        & 23.05                & \textbf{47.54}       & 31.04                & {\ul 92.91}          & \textbf{100.00}      & 66.69                & 6,171       \\
MCCE                & 17.39                & 22.98                & 19.51                & 43.79                & 84.60                & 79.11                & \textbf{1}  \\

\midrule
\multicolumn{8}{c}{Student Performance (Logistic Regression) - Min Sparsity: $38.57\%$} \\
\midrule

\textbf{L2C (Ours)} & {\ul 55.32}          & {\ul 29.54}          & {\ul 38.51}          & \textbf{100.00}      & \textbf{100.00}      & \textbf{100.00}      & {\ul 6}     \\
DICE-Random         & \textbf{87.60}       & 13.64                & 23.60                & \textbf{100.00}      & \textbf{100.00}      & {\ul 98.99}          & 2,518       \\
DICE-Genetic        & 39.20                & \textbf{39.88}       & \textbf{38.54}       & {\ul 84.83}          & \textbf{100.00}      & 60.77                & 3,406       \\
COPA                & 50.45                & 25.28                & 33.68                & 67.26                & 67.26                & 95.32                & 18,774      \\
MCCE                & 25.97                & 24.97                & 25.46                & 60.98                & {\ul 93.10}          & 67.70                & \textbf{1} 
\\
\bottomrule
\end{tabular}
}
\end{table*}

\subsection{Re-identification Risk Analysis}\label{result:privacy}

\paragraph{\textbf{Preliminaries}} We start by reviewing the fundamental concepts related to a public dataset:

\begin{itemize}[leftmargin=*]
    \item \textit{Identifiers}: Attributes that uniquely identify an individual. Identifiers can be a person's full name, government tax number or driver's license number. 

    \item \textit{Quasi-identifiers}: Attributes that themselves do not uniquely identify a person, but when combined are sufficiently correlated to at least one individual record. For example, the combination of gender, birth dates and ZIP codes can re-identify $87\%$ of American residents \cite{sweeney2000simple}. 
    
    \item \textit{Sensitive attributes}: Attribute that are protected against unauthorized access. Sensitive data is confidential and if leaked could harm personal safety or emotional well-being. Examples are salary, medical conditions, salary, criminal histories, or phone numbers.     

    \item \textit{Equivalence class}: An equivalence class is a group of records with identical quasi-identifiers.
\end{itemize}

Every public dataset must first be anonymized by removing identifiers. However, the data may still be vulnerable to re-identification attacks due to the potential existence of quasi-identifiers. To quantify the level at which a dataset is susceptible to re-identification risk, the following $3$ metrics are commonly used:

\begin{itemize}[leftmargin=*]
    \item \texttt{k-Anonymity} \cite{samarati1998protecting}: A dataset satisfies $k$-anonymity if for each record in the dataset, the quasi-identifiers are indistinguishable from at least $k-1$ other people also in the dataset.

    \item \texttt{l-Diversity} \cite{machanavajjhala2007diversity}:  A dataset has $l$-diversity if, for every equivalence class, there are at least $l$ distinct values for each sensitive attribute. 

    \item \texttt{k-Map} \cite{el2008protecting}: Given an auxiliary dataset used for re-identification (e.g., US Census or IMDb dataset in the Netflix example), so-called the `attack' dataset, a dataset satisfies $k$-map if every equivalence class is mapped to at least $k$ records in the `attack' dataset.
    
\end{itemize}

\paragraph{\textbf{Evaluation metrics}} Suppose a company releases an API that permits users to query a set of counterfactual examples. We now analyze the level of privacy leakage associated with the output data, by quantifying the percentage of successful attacks w.r.t the aforementioned metrics. Respectively, we measure (1) \texttt{1-Anonymity}: \textit{\% equivalence classes with only $k = 1$ member}, (2) \texttt{1-Diversity}: \textit{\% equivalence classes with $l = 1$ value for a sensitive attribute}, (3)  \texttt{1-Map}: \textit{\% counterfactual examples exactly matched with any single record in an `attack' dataset}. Notice that given a $k$-anonymized dataset, the existence of a one-to-one mapping with the `attack' dataset means the released dataset fails \texttt{k-Map}. 

\paragraph{\textbf{Experiments}} By definition, violations w.r.t \texttt{k-Anonymity} and \texttt{l-Diversity} are computed against the output set of examples, while \texttt{k-Map} requires an external dataset. For Adult Income, we choose the validation set as the `attack' set. For German Credit, there exists multiple versions of this dataset across the literature. The one used in our main analysis is adopted from \cite{verma2022amortized}, which has been subject to pre-processing. We use another version published by Penn State University\footnote{https://online.stat.psu.edu/stat857/node/215/} for re-identification. We assume these hold-out sets belong to some larger datasets of population availably accessed by the public. Quasi-identifiers and sensitive attributes are given in Appendix \ref{A}. Following \citet{goethals2022privacy}, we consider the black-box predicted label as part of the quasi-identifiers. 

Let's first look into the attacks on the raw output explanations presented in Table \ref{tab:privacy}. We here assume that the attacker's goal is to collect as many examples as possible without caring about which one is valid. If the attacker has no information about how the data is discretized, it is much less likely to find exact matches in the `attack', thereby reducing the re-identifiability of L2C data. Now we assume the attacker gets access to both the API and our discretization mechanism. They therefore could correspondingly discretize the data in the `attack' set and retrieve the matches. Our analysis assumes this worse scenario, meaning that for L2C only, we compute \texttt{1-Map} against the discretized data. The entries \texttt{N/A} in Table \ref{tab:privacy} are due to the fact that no counterfactual example in the data of DICE-Genetic or COPA has matches, so their robustness remains unverifiable in our experiment. We must highlight that no match does not necessarily translate to zero privacy risk. We also note that such a case is different from L2C, whose result is nearly $0.00\%$. L2C in fact still returns matches for some records wherein we achieve \texttt{k-Map} of $2-3$ specifically. Overall, L2C yields the lowest re-identifiability risk. Another interesting observation on German Credit is that although DICE performs well on \texttt{k-Anonymity}, the number of attacks on \texttt{l-Diversity} is dramatically high. This sheds light on the limitation of \texttt{k-Anonymity} discussed in \cite{machanavajjhala2007diversity} about \textit{Homogeneity attacks} and  \textit{Background knowledge attacks}. Basically, attacking a k-anonymized dataset with a high $k$ can still reveal some private information of the data subjects because profiles in the same equivalence class are similar (very few distinct values in sensitive attributes), or the adversary has some background knowledge that any help narrow down possible values. 

\begin{figure}[!h]
 \centering
  \includegraphics[width=0.80\linewidth]{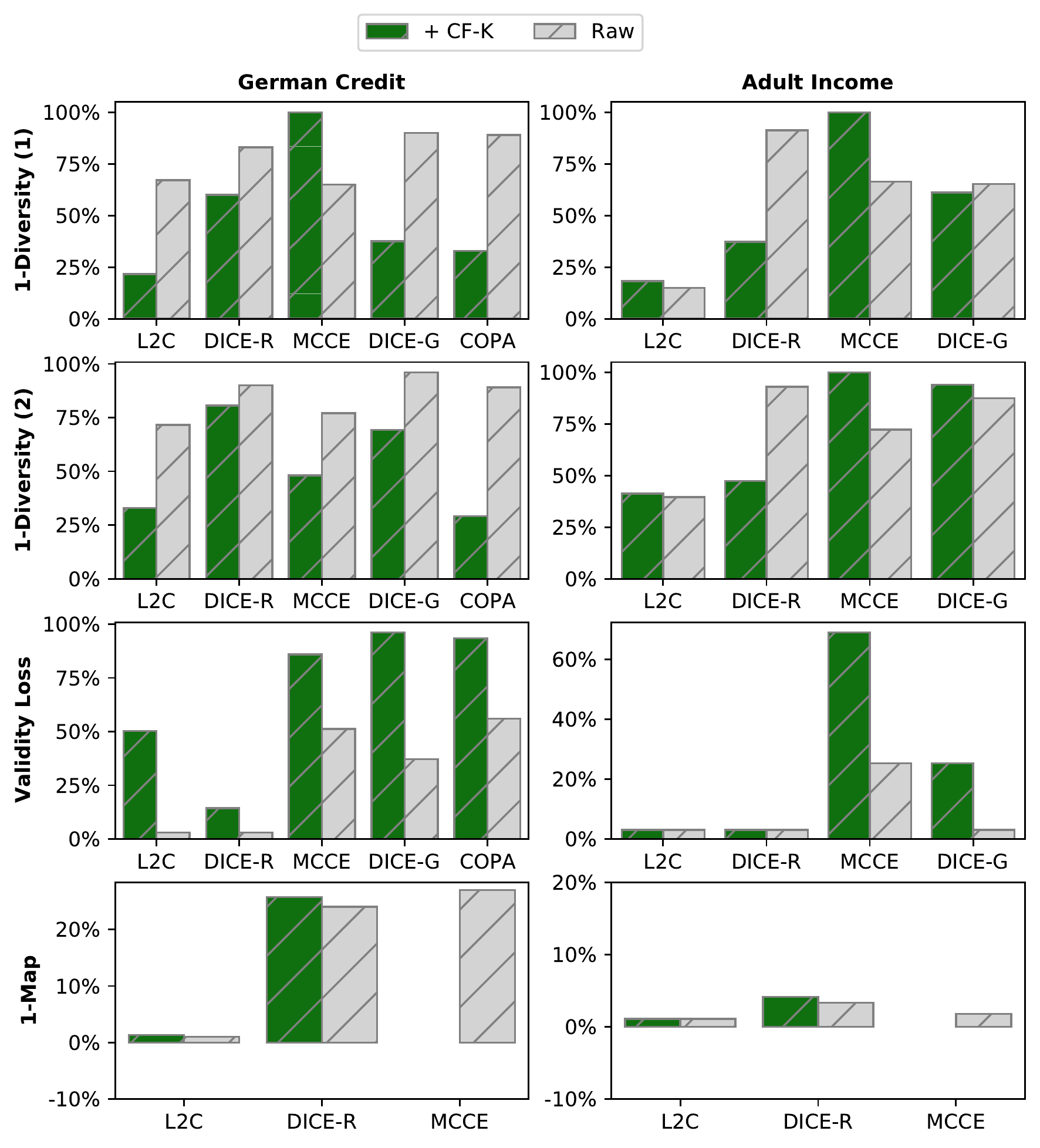}
  \caption{Privacy risk comparison between the raw output data and the data subject to $2$-anonymization under the strategy of CF-K. For all metrics, lower is better. \texttt{l-Diversity} is evaluated on $2$ most sensitive attributes.}
  \label{fig:privacy}
\end{figure}

\paragraph{\textbf{Privacy under CF-K}} We here investigate the effectiveness of the idea behind CF-K proposed in \cite{goethals2022privacy}. CF-K searches for an equivalence class for each counterfactual example and suggests only publishing profiles of at least $k$-sized equivalence class. Since the authors do not publish their codes, plus it is hugely time-consuming to run on models like DICE, we extend the above experiment and examine the effect when every output counterfactual set is $2$-anonymized. Specifically, for every set of $100$ generations, we remove records not belonging to any equivalence classes. Given now that the data is now $2$-anonymized, we evaluate attacks against \texttt{l-Diversity} and \texttt{k-Map}. We also measure \% of valid counterfactuals left in the set, assuming that a user requests for $100$ per instance. The more valid examples lost from a model explainer imply that it will be more costly to search for sufficient equivalence classes for every instance. Figure \ref{fig:privacy} depicts that in most cases, $k$-anonymization enhances the protection of the sensitive attributes but does greatly compromise the \textit{validity} of the output explanations. 

The purpose of $k$-anonymization is to ensure when attacked, an individual remains indistinguishable from at least $k-1$ others. However, notice that $k$-anonymization does not prevent \texttt{k-Map} against which the number of successful attacks is still high for DICE and MCCE. The threat is thus no less severe when the attacker is interested in the re-identifiability of both datasets. If the released data contains the sensitive information that is missing from the `attack' dataset, and given the fact that $1$-diversity remains well above zero for all methods, the attacker could easily infer private information of every linked record. To this end, generalizing the data as done in L2C is proved to be useful to prevent such an inference attack. It is also observed that combining CF-K with L2C in particular significantly improves the anonymity of our counterfactual data. We therefore believe that the integration of L2C with other privacy techniques in the cybersecurity area would yield a more effective safeguard. 

\begin{table}[!h]
\caption{Successful attacks on counterfactual explanation methods. Bold / Underline indicates the lowest / second-lowest privacy risk for each dataset. \texttt{1-Diversity} is evaluated on $2$ most sensitive attributes as shown in the columns 2 \& 3.}
\vspace{-0.5em}
\label{tab:privacy}
\resizebox{0.85\linewidth}{!}{
\begin{tabular}{l | c c c c }
\toprule
Method      & \texttt{1-Anoy.}$\downarrow$        & \multicolumn{2}{c}{\texttt{1-Diversity}$\downarrow$}  & \texttt{1-Map}$\downarrow$ 
\\

\midrule

\multicolumn{5}{c}{German Credit}       \\
\midrule

\textbf{L2C (Ours)} &  62.15\%        & {\ul 67.09\%}     & \textbf{71.60\%}     & \textbf{0.21\%}             \\
DICE-Random         & {\ul 55.15\%}              & 82.75\%              & 89.96\%              & {\ul 23.67\%}       \\
MCCE                & 62.83\%              & \textbf{ 65.40\%}        & {\ul 76.95\%}        & 26.64\%             \\
DICE-Genetic        & \textbf{15.36\%}     & 90.19\%              & 95.80\%              & N/A                  \\
COPA                & 87.61\%              & 89.41\%              & 89.24\%              & N/A              \\

\midrule
\multicolumn{5}{c}{Adult Income}       \\
\midrule

\textbf{L2C (Ours)} & \textbf{5.07\%}      & \textbf{14.90\%}     & \textbf{39.58\%}     & \textbf{0.00\%}                \\
DICE-Random         & 72.21\%              & 91.28\%              & 93.07\%              & 3.31\%                \\
MCCE                & 15.19\%              & 66.42\%              & {\ul 72.36\%}        & {\ul 1.77\%}         \\
DICE-Genetic        & {\ul 15.11\%}        & {\ul 65.32\%}        & 87.54\%              & N/A                                 
             
\\

\bottomrule
\end{tabular}
}
\end{table}
\raggedbottom

\subsection{Discretization}\label{ablation:discretization}
Discretization is an important pre-processing step in data analysis in which the problem of optimal discretization with a minimum number of splits is proved to be NP-Hard \cite{nguyen1995quantization,chlebus1998finding}. We here adopt the unsupervised \textit{Equal-frequency} discretizer, which splits a continuous attribute into buckets with the same number of instances\footnote{Except for the features Capital gain and Capital loss from Adult Income which we convert into binary variables to accurately reflect the semantics of their data.}. More concretely in this experiment, features are quantized using Python function \texttt{qcut}\footnote{\url{https://pandas.pydata.org/docs/reference/api/pandas.qcut.html}}, which requires specifying the maximum number of buckets/levels and later adjusts it depending on the input data distribution. We set the maximum buckets to be $4$ such that every bucket averagely has $25\%$ of total observations. Table \ref{tab:discretization_quali} reports how the numerical features of each dataset are discretized. We argue that having very few buckets is likely to cause under-fitting since there are very few useful combinatorial patterns that can counter the original label. Whereas we need diversity for effective learning, too many buckets are undesirable since it can hurt generalization due to some following reasons : (1) each bucket would contain too little data and the chosen middle value may not represent the bucket well, and (2) the model has more combinations of features to explore, thus can converge to sub-optimal combinations that cannot generalize well on unseen test points. In this regard, we decide to split data into equal-sized buckets in the hope of balancing the trade-off.

Various discretization methods exist \cite{ramirez2016data}. Table \ref{tab:discretization_quanti} analyzes the performance of our method L2C under $3$ other discretization strategies. The first one is \textit{Minimal entropy partitioning (MDP)} \cite{fayyad1993multi}. It is an old-school supervised approach and one of the most widely used. MDP determines the binary discretization for a value range by selecting the cut point that minimizes the class entropy. The algorithm can be applied recursively on sub-partitions induced by the initial cut point and the paper proposes using Minimum Description Length Principle\footnote{\url{https://orange3.readthedocs.io/}} \cite{rissanen1978modeling} as a stopping condition. Another strategy is to apply \textit{Domain knowledge} where feature values can be grouped based on common demographic or social characteristics. For example, Age could be translated into different age groups (e.g., Teenagers, Young Adults, etc.), or TOEFL scores are divided into proficiency levels. While the aforementioned methods are uni-variate, we also implement a multivariate approach using Decision tree. Following the motivation of MDP, we run \textit{CART} \cite{breiman2017classification} to search for the splits that minimize class information entropy. Note that we here consider the predicted labels from the black-box models as the target variable. The goal is to each sure the prediction on each combination of features is stable as possible. To avoid fine-grained intervals, we set the minimum number of samples for a split to be $30$. Our experiment shows that we are still able to achieve $100\%$ of \texttt{Validity} and \texttt{Coverage} in roughly the same amount of time. We therefore only present the remaining metrics in Table \ref{tab:discretization_quanti}, which here demonstrates a comparable quality of explanations among different discretization strategies. Given that L2C performance is relatively insensitive to the choice of discretizers, we therefore suggest using \textit{Equal-frequency} for which no labels or external knowledge is required.


\begin{table}[h]
\caption{Desirability of L2C counterfactual explanations under various discretization strategies. $^*$Proposed method.}
\vspace{-0.5em}
\label{tab:discretization_quanti}
\resizebox{\linewidth}{!}{
\begin{tabular}{l |p{1.5cm} p{1.5cm} p{1.5cm} p{2cm} p{2cm} }
\toprule
Strategy  &  \texttt{Sparsity (\%)}$\uparrow$ & \texttt{Diversity (\%)}$\uparrow$ &  \texttt{Harmonic Mean (\%)}$\uparrow$  & \texttt{Unary} (\%)$\uparrow$  \\
\midrule
\multicolumn{5}{c}{German Credit - Min Sparsity: $20.00 \%$} \\
\midrule

Equal Freq.$^*$ & 61.35 & 37.31 & 46.39 & 99.06\\
MDP & 61.58 & 35.00 & 44.63 & 100.00\\
CART & 60.92 & 39.85 & 48.18 & 100.00\\
Domain Know. & 61.73 & 37.56 & 46.70 &  100.00\\

\midrule
\multicolumn{5}{c}{Graduate Admission - Min Sparsity: $14.29 \%$} \\
\midrule

Equal Freq.$^*$ & 42.23 & 37.90 & 39.94 & 100.00\\
MDP & 58.20 & 41.34 & 41.56 & 100.00\\
CART & 41.80 & 41.34 & 41.56 & 100.00\\
Domain Know. & 42.17 & 42.30 & 42.22 & 100.00\\

\midrule
\multicolumn{5}{c}{Student Performance - Min Sparsity: $38.57 \%$} \\
\midrule
Equal Freq.$^*$ & 55.32 & 29.54 & 38.51 & 100.00\\
MDP & 55.57 & 28.50 & 37.67 & 100.00\\
CART & 55.07 & 27.79 & 36.94 & 100.00\\
Domain Know. & 55.47 & 31.57 & 40.23 & 100.00\\

\bottomrule
\end{tabular}
}
\end{table}

\subsection{Feature Correlational Constraints}\label{binary_constraint}
While the treatment of unary constraints is straightforward for heterogeneous datasets, we argue that this is not the case for binary constraints. For example, suggesting that a person get a Master's degree at precisely the age of $34$ is unrealistically rigid. This issue indeed stems from the presence of continuous features. A direct solution is to allow for more flexible suggestions through discretization (e.g., suggesting an age range from $30-40$ instead of an exact value at $34$). This indeed aligns with the generative mechanism of our L2C, which sets us apart from existing works. However, discretization is currently treated as a subroutine of internal processing, meaning that in the output examples, the values for the continuous features are still returned in the numerical format for the sake of consistency. Therefore, the best strategy would be to have the machine learning classifiers trained in the discretized feature space accordingly. Practitioners could then ignore the one-hot decoding stage and deploying L2C for this purpose would be effortless. In Appendix \ref{constraint}, we demonstrate how L2C effectively addresses binary constraints in this scenario with success rates of 91.54\% and 100.00\% respectively on German Credit and Adult Income.

\section{Conclusion and Future Work}\label{conclusion}
In this paper, we study the challenges facing algorithmic recourse approaches in generating diverse counterfactual explanations: how \textit{diversity} can be tackled without compromising the other desiderata of an explanation while preserving privacy against linkage attacks. We analyze how existing engines fail to resolve the trade-offs among counterfactual constraints and fill the research gap with our novel framework \textbf{L2C}. Here we target a broad class of differentiable machine learning classifiers. To fit non-differentiable models in our framework, one could use policy gradient \citep{sutton1999policy} or attempt to approximate such models as decision trees or random forests with a differentiable version \citep{yang2018deep,lucic2022focus}. L2C is currently proposed to deal with re-identification risks of the released examples. Defense against \textit{model stealing} and \textit{membership inference attacks} however remains exigent. Integrating differential privacy in the framework of L2C is one interesting research avenue, which we leave for future works to explore.  

\section{Acknowledgement}
Dinh Phung and Trung Le gratefully acknowledge the support by the US Airforce FA2386-21-1-4049 grant and the Australian Research Council ARC DP230101176 project. This does not imply endorsement by the funding agency of the research findings or conclusions. Any errors or misinterpretations in this paper are the sole responsibility of the authors.

\bibliographystyle{ACM-Reference-Format}
\bibliography{bibliography}

\clearpage
\appendix

\section{Experimental Details}\label{A}
\subsection{Dataset statistics}

\begin{itemize}[leftmargin=*]
    \item \textbf{German Credit} \citep{Dua:2019}: This dataset includes information of customers taking credit at a bank. The task is to classify a customer as a good (label $1$) or bad (label $0$) credit risk. 
    
    \item \textbf{Adult Income} \citep{kohavi1996scaling}: The dataset was extracted from the US 1994 census data on adult incomes. The task is to classify if an individual's income exceeds \$50,000 per year (label $1$) or not (label $0$). 
    
    \item \textbf{Graduate Admission} \citep{acharya2019comparison}: The set contains data of Indian students' applications to a Master's program. The original target variable is an ordinal variable on the scale of $[0-1]$ indicating the chance of a student being admitted, where $1$ indicates the highest chance. We set a threshold of $0.7$ and re-categorize students as either "having a higher chance" ($ \ge 0.70$-label $1$) or "having a lower chance" ($ < 0.70$-label $0$). The binary classification task is to determine if a student profile has a higher chance of being successful at their application. 
    
    \item \textbf{Student Performance} \citep{cortez2008using}: This dataset records the performance of students at two schools Gabriel Pereira and Mousinho da Silveira. The task is to predict if a student achieves a final score above average (label $1$) or not (label $0$). The train and test splits contain data of students from these two schools separately \citep{bui2022counterfactual}.

\end{itemize}

Table \ref{tab:datastats} summarizes the experimental settings for every dataset. Regarding the underlying black-box models, we experiment with Logistic Regression for the linear classifier and Neural Network for the non-linear classifier. We train linear classifiers on German Credit and Student Performance, and non-linear classifiers on Graduate Admission (with $3$ layers and $40$-dimensional hidden units) and Adult Income (with $3$ layers and $30$-dimensional hidden units). For each task, we further sample $20\%$ random observations of the training sets as validation sets and train $5$ black-box models with the same architecture but with different initializations. Since we have specified a fixed subset of immutable features, we therefore can derive the minimum sparsity level an explanation method should obey as 
$$\textrm{Min Sparsity} = \frac{\textrm{No. immutable features}}{\textrm{Total no. features}}$$

\subsection{Model Design}
We parameterize $\mathcal{G}$ and $\mathcal{S}$ with neural networks of $3$ and $2$ layers respectively. Each layer consists of a dense layer and a ReLU activation, except the last layer of $\mathcal{S}$ takes Sigmoid activation to produce a probability vector. The final layer of $\mathcal{G}$ is another dense layer that outputs a logit vector of the same dimension as the input, representing the counterfactual distribution. We set the sparsity loss coefficient $\alpha = 1e-4$ and use the same architecture for all tasks. We train our model with Adam optimizer for $200$ epochs, at $\tau = 0.2$ and a learning rate of $1e-4$.  

\section{Feature Constraints of Discretized Black-box Models} \label{constraint}

We here demonstrate how binary constraints can be effectively addressed within L2C framework for realistic explanations. We use the same model architecture and experimental setup, except that the data now is initially discretized and the black-box classifiers are trained on the discretized feature space. Not only would this help produce more realistic suggestions but it would also enhance privacy protection. We reuse the \textit{Equal-frequency} strategy for consistency, yet note that practitioners are highly recommended to consider generalizing the data in such a way that satisfies the privacy metrics outlined in Section \ref{result:privacy}. We then run L2C models on German Credit and Adult Income datasets with feature correlations specified in Table \ref{tab:datastats}: respectively the constraints \textit{Present residence} $\rightarrow$ \textit{Age} and  \textit{Education level} $\rightarrow$ \textit{Age}. To enforce the increasing constraint on the child variable (i.e. \textit{Age}), we further track the perturbation of the parent variable (i.e. \textit{Present residence}, \textit{Education level}) and apply the Eq. (\ref{eq:constraint}) to update the distribution of samples where the parent variable is indicated (by generator $\mathcal{G}$) to increase. Table \ref{tab:constraint} reports the quality of explanations produced under both types of constraints, where \texttt{Binary} measures the proportion of examples $\mathbb{C}$ meeting the binary constraints. Here we again substantiate the effectiveness of L2C in balancing counterfactual constraints.    
 
\begin{table}
\caption{Desirability of counterfactual examples generated from L2C for explaining a discretized black-box classifiers.}
\vspace{-0.5em}
\label{tab:constraint}
\resizebox{\linewidth}{!}{
\begin{tabular}{p{1.5cm} p{1.5cm} p{1.5cm} p{1.5cm}}
\toprule
\texttt{Harmonic Mean (\%)}$\uparrow$ &  \texttt{Validity (\%)}$\uparrow$ & \texttt{Unary (\%)}$\uparrow$ & \texttt{Binary (\%)}$\uparrow$  \\
\midrule
\multicolumn{4}{c}{German Credit} \\
\midrule
48.56\% & 100.00\% & 98.38\% & 91.54\% \\
\midrule
\multicolumn{4}{c}{Adult Income} \\
\midrule
23.89\% & 100.00\% & 100.00\% & 100.00\% \\

\bottomrule
\end{tabular}
}
\end{table}

\begin{table*}[!h]
\caption{Dataset statistics. $^*$Features subject to non-decreasing constraints.}
\vspace{-0.5em}
\label{tab:datastats}
\centering
\resizebox{\textwidth}{!}{
\begin{tabular}{p{3cm}|p{3.5cm}|p{3.5cm}|p{2.5cm}|p{2.5cm}}
\toprule
Dataset & \textbf{German} & \textbf{Adult} & \textbf{Graduate} & \textbf{Student} \\
 & \textbf{Credit} & \textbf{Income} & \textbf{Admission} & \textbf{Performance} \\
\toprule
 Train/Dev/Test & $640/160/200$ & $28942/7235/9045$ & $320/80/100$ & $339/84/226$  \\ 
 No. features & $20$ & $13$  & $7$ & $14$ \\ 
 \midrule

 \multirow{4}{*}{Immutable features} & Foreign worker & Race & University  & Mother's edu. \\ 
 
 & No. liable people & Sex & rating & Father's edu. \\ 
 
 & Personal status & Native country & & Family edu. support \\
 & Purpose & Marital status & & First period grade \\
 \midrule

 \multirow{4}{*}{Feature constraints$^*$} & Age & Age & Research & Age \\ 
 & Present employment & Education level & experience & \\
 & Present residence & & & \\
 & Duration & & & \\
 \midrule

 Feature correlations & Increasing Age increases Present residence  & Increasing Education level increases Age & & \\  
 \midrule

 \multirow{6}{*}{Quasi-identifiers} & Age, Job & Age & & \\ 
 & Foreign worker & Sex & & \\
 & Personal status & Race & & \\
 & Present employment & Relationship & & \\
 & Present residence & Marital status & & \\
 & Property, Housing & & & \\
 \midrule

 \multirow{2}{*}{Sensitive features} & Credit amount & Capital gain & & \\ 
 & Savings account & Capital loss & & \\
 \midrule
 
 Black-box model & Logistic Regression & Neural Network & Neural Network & Logistic Regression \\

 Test accuracy & $67.00\%$ & $85.53\%$ & $90.60\%$ & $94.69\%$ \\ 

 Min Sparsity & $20.00\%$ & $30.77\%$ & $14.29\%$ & $38.57\%$ \\
\bottomrule
\end{tabular}
}
\end{table*}

\section{Amortized Baselines}\label{ablation:amortized}
Table \ref{tab:amortized} compares L2C with popular amortized approaches: Feasible-VAE \citep{mahajan2019preserving}, CRUDS \citep{downs2020cruds} and FastAR \citep{verma2022amortized} across desiderata. We now provide the experimental setup for the amortized baseline models. Whereas the non-amortized algorithms are run directly on the testing sets, for amortized methods, we train the base generative models on the training sets and use the testing sets only for evaluation. We tune the base generative models under various different hyper-parameter settings via grid search and report the best results. We determine the best settings via two metrics: \texttt{Coverage} and \texttt{Diversity}. When there is a trade-off, \texttt{Coverage} is chosen to be the deciding criterion. 

Specifically, for Feasible-VAE \citep{mahajan2019preserving}, we tune the hidden dimensions of the VAE encoder within $\{10, 30, 50, 70, 90\}$ and regularization term on \textit{Validity} within $\{42, 62, 82, 102, 122\}$. For CRUDS \citep{downs2020cruds}, the base model is a Conditional Subspace Variational Auto-encoder \citep{klys2018learning}. In the original paper, the network only has $1$ hidden layer of $64$ nodes, which we find to be of low capacity. We thus experiment with $2$ layers and different hidden dimensions within $\{16, 32, 64\}$. For FastAR \citep{verma2022amortized}, the hyper-parameters include manifold distance $\lambda$ and entropy loss coefficient. Across datasets, \citet{verma2022amortized} shows the best \texttt{Coverage} under $\lambda = 0.1$. We thus set $\lambda = 0.1$ as well in our experiments, while focusing on tuning the latter hyper-parameter within $\{0.01, 0.05, 0.1, 0.5, 1.0\}$. For the remaining hyper-parameters, we adopt the best values reported by the authors. A pre-trained FastAR model can only interpret one decision outcome chosen as the desired one (often the positive label). We must therefore train separate FastAR models on the positive and negative subsets and combine the results. We further find that although it is straightforward to obtain multiple generations in a single model, FastAR algorithm is optimized for one optimal counterfactual state for a given input. Thus in the hope of achieving better \textit{diversity}, we train $100$ different model initializations and accordingly collect a set of $100$ explanations for evaluation. The inference time accumulates as a result, which is the reason why our reported results on time efficiency for FastAR are different from what are reported in the authors' paper.

\section{The Role of Feature Selector}\label{ablation:selector}

\begin{figure}[!h]
 \centering
  \includegraphics[width=0.35\textwidth]{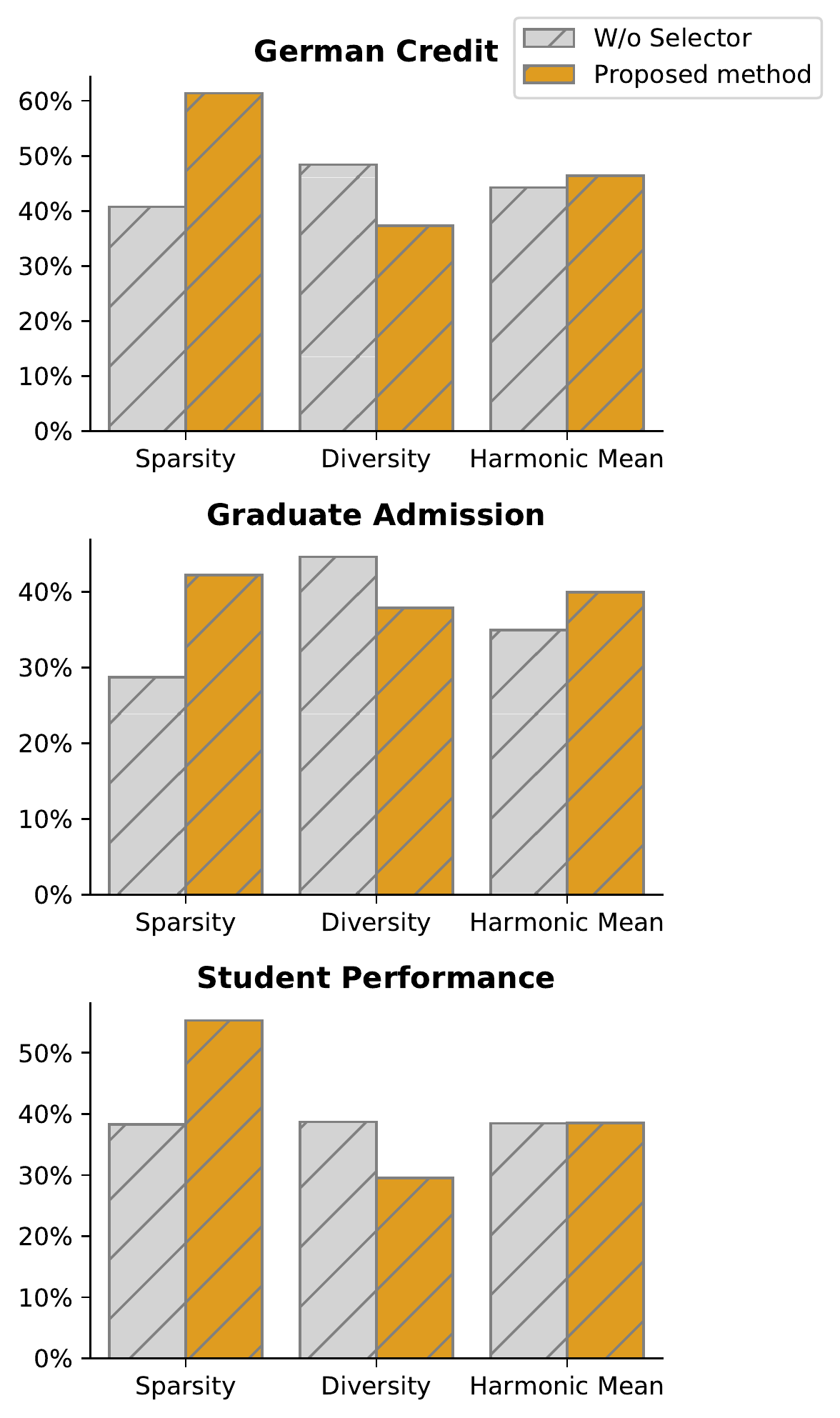}
  \caption{Analysis of L2C performance when the selector $\mathcal{S}$ is removed.}
  \label{fig:selector}
\end{figure}

We now validate the importance of learning the local feature-based selection distribution via the feature selector $\mathcal{S}$. We 
 first remove $\mathcal{S}$ from L2C framework and replace the probability vector $\boldsymbol{\pi}(\vz)$ with a binary mask vector $\vm \in [0,1]^N$ where $m_i = 1$ if $i \in \mathbb{K}$ (i.e., a mutable feature) and $m_i = 0$ otherwise. We thus use $m_i$ in substitution of $s_i$ to update the counterfactual representations $\widetilde{\vz_i}$ as previously done. We only optimize the generator $\mathcal{G}$ to learn the feature-based perturbation distribution, and the training objective Eq. (\ref{eq:new_obj}) excludes the regularization term for \textit{sparsity} accordingly. Figure \ref{fig:selector} investigates the performance of L2C under this alternative setup, in comparison with the proposed method that jointly optimizes $\mathcal{S}$ and $\mathcal{G}$. L2C still achieves $100\%$ of \texttt{Validity} and \texttt{Coverage}, so we only report the relevant metrics.

One drawback of omitting the Selector component is that we lose the flexibility in tailoring the quality of counterfactual generations to potential user preferences. It is seen that the Selector introduces significant \textit{sparsity} to gain an effective balance for the trade-off against \textit{diversity}. Furthermore, these results support our claim about the role of the generator $\mathcal{G}$ in that the perturbation distribution alone can yield impressively diverse explanations with \textit{sparsity} remaining under the required maximum level. This is a benefit of learning an entire feature distribution in that sometimes an output sample falls into the original input value i.e., $\widetilde{\vz_i} = \vz_i$ while combining adequately with other features.

\section{Qualitative Examples}\label{quali}

Table \ref{tab:quali_german} - \ref{tab:quali_student} illustrate some examples of our generated counterfactuals for each dataset. For illustration purposes only, we report the discretized values for numerical features where the edge values of each numerical interval are rounded to the nearest whole number. Immutable features are italicized.

\begin{table*}[!h]
\caption{Desirability of amortized counterfactual explanation methods. $\downarrow$ Lower is better. $\uparrow$ Higher is better. Bold / Underline indicates the best / second-best performance for each dataset. \texttt{Time} records total inference time in seconds.}
\vspace{-0.5em}
\label{tab:amortized}
\resizebox{\textwidth}{!}{
\begin{tabular}{l|c c c c c c r }
\toprule
Method  &  \texttt{Sparsity (\%)}$\uparrow$ & \texttt{Diversity (\%)}$\uparrow$ &  \texttt{Harmonic Mean (\%)}$\uparrow$ & \texttt{Validity (\%)}$\uparrow$ & \texttt{Coverage (\%)}$\uparrow$ & \texttt{Unary} (\%)$\uparrow$ & \texttt{Time(s)}$\downarrow$ \\
\midrule
\multicolumn{8}{c}{German Credit (Logistic Regression) - Min Sparsity: $20.00 \%$} \\
\midrule
\textbf{L2C (Ours)} & {\ul 61.35}                & \textbf{37.31}       & \textbf{46.39}       & \textbf{100.00}      & \textbf{100.00}      & \underline{99.06}       & \textbf{18}                   \\
FastAR & \textbf{95.93}  & 0.68  & 1.33  & {\ul 95.79}  & {\ul 95.79}  & \textbf{99.87}  & 10,605 \\
F-VAE  & 45.93 & 1.59  & 3.06  & \textbf{100.00} & \textbf{100.00} & 74.41  & {\ul 36}  \\
CRUDS  & 29.72 & {\ul 14.21} & {\ul 19.18} & 60.00  & 60.00  & 71.31  & 42,920 \\

\midrule
\multicolumn{8}{c}{Graduate Admission (Neural Network) - Min Sparsity: $14.29 \%$} \\
\midrule

\textbf{L2C (Ours)} & \textbf{42.23}    & \textbf{37.90}       &  \textbf{39.94}       & \textbf{100.00}      & \textbf{100.00}      & \textbf{100.00}      & \textbf{4}           \\
FastAR & {\ul 27.97} & 1.32  & 2.35  & {\ul 87.41}  & {\ul 87.41}  & {\ul 85.28}  & 5,405  \\
F-VAE  & 7.71 & 1.48  & 1.84  & \textbf{100.00} & \textbf{100.00} & 44.60  & {\ul 16}     \\
CRUDS  & 11.31 & {\ul 17.38} & {\ul 13.40} & 60.28  & 76.00  & 49.00  & 21,460 \\

\midrule
\multicolumn{8}{c}{Student Performance (Logistic Regression) - Min Sparsity: $38.57 \%$} \\
\midrule
\textbf{L2C (Ours)} & {\ul 55.32}                & {\ul 29.54}       & \textbf{38.51}       & \textbf{100.00}      & \textbf{100.00}      & \textbf{100.00}      & \textbf{6}           \\
FastAR & \textbf{81.86}  & 1.44  & 2.77  & {\ul 97.71}  & 97.71  & {\ul 99.76}  & 16,370 \\
F-VAE  & 27.37 & 7.68  & 11.92 & \textbf{100.00} & \textbf{100.00} & 69.68  & {\ul 36}     \\
CRUDS  & 24.06 & \textbf{33.64} & {\ul 28.55} & 62.12  & \textbf{100.00} & \textbf{100.00} & 39,571 \\

\bottomrule
\end{tabular}
}
\end{table*}

\raggedbottom

\section{Additional Analysis on Discretizers}
In the main paper, we have proven that the quality of L2C explanations is relatively insensitive to the choice of discretizers. This means that we can achieve interpretability without compromising the desiderata of a counterfactual explanation. Here we conduct an additional privacy analysis for similar purpose. In Table \ref{tab:privacy_discretizer}, we report our privacy analysis results on German dataset across different discretization strategies. We reuse the settings reported in the paper. It is observed that there is a same pattern with the desiderata analysis: regardless of the choice of discretizers, the performance of L2C compared to the baseline methods remains relatively stable. 

\begin{table}[!h]
\caption{Privacy of L2C counterfactual explanations under various discretization strategies on German dataset. $^*$Proposed method.}
\vspace{-0.5em}
\label{tab:privacy_discretizer}
\resizebox{\linewidth}{!}{
\begin{tabular}{l|llll}
\toprule
Strategy     & \texttt{1-Anoy.} $\downarrow$ & \multicolumn{2}{l}{1-Diversity $\downarrow$} & \texttt{1-Map} $\downarrow$ \\
\midrule
Equal Freq.$^*$   & 62.15\%              & 67.09\%               & 71.60\%              & 0.21\%             \\
MDP          & 59.03\%              & 66.52\%               & 68.40\%              & 0.10\%             \\
CART         & 65.88\%              & 69.24\%               & 74.25\%              & 0.18\%             \\
Domain Know. & 60.85\%              & 65.84\%               & 70.07\%              & 0.23\%   
\\         
\bottomrule
\end{tabular}
}
\end{table}

For L2C, it is also possible to leverage different discretization methods on different features. In practice, this strategy is highly recommended based on user demand and/or human knowledge. In Tables \ref{tab:mixed_result} and \ref{tab:mixed_privacy} , we report the performance of L2C on Adult dataset under Mixed discretization strategy. Specifically, we randomly assign one of the four mentioned strategies (Equal Frequency/MDP/CART/Domain Knowledge) to every continuous feature. The results are averaged over $5$ different random initializations. It is seen that the performance of L2C is well above the baselines where the privacy level remains rather stable again.

\begin{table*}
    \centering
    \caption{Desirability of L2C counterfactual explanations under mixed discretization on Adult dataset. $^*$Proposed method.}
    \label{tab:mixed_result}
    \resizebox{\linewidth}{!}{
    \begin{tabular}{l|c c c c c c r }
    \toprule
 Strategy  & \texttt{Sparsity} (\%)$\uparrow$ & \texttt{Diversity} (\%)$\uparrow$ &  \texttt{Harmonic Mean} (\%)$\uparrow$ & \texttt{Validity} (\%)$\uparrow$ & \texttt{Coverage} (\%)$\uparrow$ & Unary (\%)$\uparrow$ & \texttt{Time} (s)$\downarrow$ \\
    \midrule
    Equal Freq.$^*$ & 45.70          & 28.11      & 34.80       & 100.00      & 100.00      & 97.62       & 444   \\

    Mixed & 43.62 &	38.83 &	41.00 & 	100.00	& 100.00	& 100.00 &	378 \\
    \bottomrule
    \end{tabular}
    }
\end{table*}

\begin{table}
\caption{Privacy of L2C counterfactual explanations under mixed discretization on Adult dataset. $^*$Proposed method.}
    \label{tab:mixed_privacy}
\begin{tabular}{l|cccc}
\toprule
Strategy     & \texttt{1-Anoy.} $\downarrow$ & \multicolumn{2}{l}{\texttt{1-Diversity} $\downarrow$} & \texttt{1-Map} $\downarrow$ \\
\midrule
Equal Freq.$^*$   & 5.07\%	& 14.90\%	& 39.58\% &	0.00\%         \\
Mixed          & 5.53\%	& 15.39\% &	38.43\%	& 0.00\%           \\
\bottomrule
\end{tabular}
\end{table}

\section{Degree of Feature Mutability}

Following the setting from previous works, we currently limit our discussion to binary states of mutability (i.e., whether a feature gets changed or not). However, in practice, actionability should further deal with the difficulty level associated with mutable features, also among their categorical levels. Although this extension remains beyond existing works and ours, it can be seen that L2C can be flexibly modified to handle these scenarios. We here sketch some ideas for future works to explore.

Based on human experts, one can define a (constant) cost vector $\phi \in [0,1]^K$ that reflects the (relative) level of difficulty among mutable features (where $K$ is the number of mutable features and $\sum^{K}_{i=1} \phi_i = 1$). Recall in Section 3.3 that the feature-based selection distribution (i.e., Bernoulli($\pi_i \vert z$)) for every feature $i$ is given as 
$$ \pi_i(z) = \frac{1}{1 + \exp{\big\{-\mathcal{S}_i(z)}\big\}}.$$

We want the features with high-cost values to be sampled less likely, meaning to have a low probability of being changed. To lower such a probability, while facilitating differentiable training, for every feature $i$, one can re-define the feature-based selection distribution as $(1 - \phi_i) \times \pi_i(z)$, or 
$$
\pi_i(z) =  (1 - \phi_i) \times \frac{1}{1 + \exp{\big\{-\mathcal{S}_i(z)}\big\}}.
$$

Another cost vector can also be defined for each categorical level of a feature. One could reuse our mechanism to handle Unary constraints in Equation (5) to explicitly force the model to generate more samples at the low-cost levels (See Lines 494 - 520). 
Let $\phi_{ij}$ denote the cost of category $j$ of feature $i$, the perturbation distribution can be re-defined as 
$$
p_{ij}(z) \propto (1-\phi_{ij}) \times \frac{ \exp \big\{ \mathcal{G}_{ij}(z) \big\}}  {\sum^{c_i}_{k=1} \exp \big\{ \mathcal{G}_{ik}(z) \big\} }, \forall j = 1,...,c_i,$$

\section{L2C under high dimensionality}

Aware of the fact that high-dimensionality and sparsity can affect any systems' performance, we conduct an additional analysis to investigate the performance of L2C on a high-dimensional textual dataset. We choose the use case of Email Spam Detection\footnote{\url{https://archive.ics.uci.edu/ml/datasets/spambase}} where the feature set is the vocabulary of the entire corpus and each word is considered a binary feature (indicating whether a word appears in an email or not). A possible scenario is an attacker somehow gains access to the detector and would like to build a model explainer that can tell which word should be removed in order to fool the detector. 

This dataset contains up to $2,136$ features (after flattening), thus being extremely sparse. To make it comparable, we reuse the same model architecture and training settings in the other datasets. Table \ref{tab:val_eff} reports the Validity and Inference Time (per input when generating a set of $100$ samples) of L2C explanations on this dataset, compared with the other datasets of smaller scale.  

\begin{table}
\caption{Validity and Time Efficiency of L2C w.r.t feature size.}
\label{tab:val_eff}
\resizebox{\linewidth}{!}{
\begin{tabular}{l|ccc}
\toprule
Dataset             & No. Features & \texttt{Validity} & Time per input \\
\midrule
Graduate Admission  & 34           & 100\%    & 40 ms           \\
Student Performance & 48           & 100\%    & 30 ms          \\
German Credit       & 78           & 100\%    & 90 ms          \\
Adult Income        & 109          & 100\%    & 50 ms          \\
Spam Detection      & 2,136        & 87\%     & 13 s        \\
\bottomrule
\end{tabular}
}
\end{table}

We observe that for some extremely sparse data points, it is indeed difficult to find enough valid counterfactuals within the time budget of 5 minutes, leading to a total decrease in validity and an increase in the average inference time, compared to the smaller datasets. However, even for such a challenging dataset, L2C still works on a large number of inputs to achieve an overall accuracy of $87\%$. If we compare L2C with DiCE - the closest baseline in performance (also the most popular), we posit that L2C is more scalable since DiCE iteratively perturbs every single feature, hence would be catastrophically expensive when dealing with text data. Furthermore, we argue that if the dataset contains more of the categorical features, it poses the same serious issue to all counterfactual generation systems. Sparsity and curse of dimensionality are thus long-standing problems in this research area, which would result in a dedicated paper altogether. To address it may require more careful investigation into the model design and hyper-parameter tuning, which we leave open for future works.

\clearpage


\begin{table*}[hbt!]
    \caption{Discretization of numerical features in each dataset.}
    \label{tab:discretization_quali}
    \vspace{-0.5em}
    \centering
    \resizebox{0.45\linewidth}{!}{
    \begin{tabular}{l | c | r }
    \toprule
    \textbf{Feature name} & \textbf{Bucket values} & \textbf{No. Data points}  \\ 
    \toprule
    \multicolumn{3}{c}{German Credit} \\ 
    \midrule
    \multirow{3}{*}{Duration (Months)} & ( 4 , 12 ] & 359 \\
    & ( 12 , 18 ] & 187 \\
    & ( 18 , 24 ] & 224 \\
    & ( 24 , 72 ] & 230 \\
    \midrule
    \multirow{3}{*}{Credit Amount} & ( 247 , 1367 ] & 250 \\
    & ( 1367 , 2320 ] & 250 \\
    & ( 2320 , 3971 ] & 250 \\
    & ( 3971 , 18425 ] & 250 \\
    \midrule
    \multirow{3}{*}{Age} & ( 19 , 27 ] & 291 \\
    & ( 27 , 33 ] & 225 \\
    & ( 33 , 42 ] & 249 \\
    & ( 42 , 75 ] & 235 \\

    \toprule
    \multicolumn{3}{c}{Adult Income} \\ 
    \midrule
    \multirow{3}{*}{Age} & ( 18 , 30 ] & 14,260 \\
    & ( 30 , 45 ] & 17,727 \\
    & ( 45 , 60 ] & 10,387 \\
    & ( 60 , 75 ] & 2,848 \\
    \midrule
    Capital gain & $\le$ 0 & 41,432 \\
    & > 0 & 3,790 \\
    \midrule
    Capital loss & $\le$ 0 & 43,082 \\
    & > 0 & 2,140 \\
    \midrule
    Hours per week & ( 0 , 20 ] & 3,602 \\
    &   ( 20 , 40 ] & 27,843 \\
    & ( 40 , 100 ] & 13,777 \\
    
    \toprule
    \multicolumn{3}{c}{Graduate Admission} \\ 
    \midrule
    \multirow{3}{*}{GRE score} & ( 290 , 312 ] & 189 \\
    & ( 312 , 322 ] & 148 \\
    & ( 322 , 340 ] & 163 \\
        \midrule
    \multirow{3}{*}{TOEFL score} &  ( 92 , 104 ] & 176 \\
    & ( 104 , 110 ] & 175 \\
    & ( 110 , 120 ] & 149 \\
        \midrule
    \multirow{3}{*}{Undergraduate GPA} & ( 7 , 8 ] & 170 \\
    & ( 8 , 9 ] & 163 \\
    & ( 9 , 10 ] & 167 \\
    
    \toprule
    \multicolumn{3}{c}{Student Performance} \\ 
    \midrule
    \multirow{3}{*}{Age} & ( 15 , 16 ] & 289 \\
    & ( 16 , 17 ] & 179 \\
    & ( 17 , 22 ] & 181 \\
    \midrule
    \multirow{2}{*}{School absences} & ( -0 , 4 ] & 466 \\
    & ( 4 , 32 ] & 183 \\       
    \midrule 
    \multirow{3}{*}{First period grade} & ( -0 , 10 ] & 252 \\
    & ( 10 , 13 ] & 245 \\
    & ( 13 , 19 ] & 152 \\
     \midrule
    \multirow{4}{*}{Second period grade} & ( -0 , 10 ] & 228 \\
    & ( 10 , 13 ] & 269 \\
    & ( 13 , 19 ] & 152 \\

    \bottomrule
    \end{tabular}
    }
    
\end{table*}

\begin{table*}[!hbt]
\caption{Counterfactual examples from German Credit dataset. *DM: Deutsche Mark}
\centering
\label{tab:quali_german}
\vspace{-0.5em}
\resizebox{\textwidth}{!}{
\begin{tabular}{l| r | r | r | r | r | r}
\toprule
\textbf{German Credit}        & Original input  & \multicolumn{5}{c}{Counterfactuals} \\
        &  (Bad credit risk) & \multicolumn{5}{c}{(Good credit risk)} \\
\midrule
Duration (months)                & 12 - 18              & 24 - 72              & 18 - 24              & 12 - 18              & 18 - 24              & 12 - 18              \\ [1ex]
Credit amount (DM)               & 1367 - 2320          & 1367 - 2320          & 1367 - 2320          & 1367 - 2320          & 247 - 1367           & 247 - 1367           \\[1ex]
Age                              & 19 - 27              & 19 - 27              & 19 - 27              & 19 - 27              & 19 - 27              & 42 - 75              \\[1ex]
Checking account (DM)            & no account           & 200+                 & no account           & 200+                 & under 200            & Under 0              \\[1ex]
Credit history                   & paid back duly       & no credit taken      & paid back duly       & paid back duly       & paid back duly       & other credits        \\[1ex]
\textit{Purpose}                 & furniture            & furniture            & furniture            & furniture            & furniture            & furniture            \\[1ex]
Savings account (DM)             & under 100            & under 500            & 1000+                & 1000+                & under 100            & under 100            \\[1ex]
Present employment since         & 1 - 4                & 1 - 4                & 1 - 4                & 7+                   & 1 - 4                & 1 - 4                \\ [1ex]
Installment rate                 & Under 20             & Under 20             & Under 20             & Under 20             & Under 20             & Under 20             \\ [1ex]

\multirow{2}{*}{\textit{Personal status}}          & Male   & Male   & Male   & Male   & Male   & Male   \\ [1ex]
& divorce   & divorced   & divorced   & divorced   & divorced   & divorced   \\ [1ex]
Other debtors                    & none                 & none                 & guarantor            & none                 & none                 & guarantor            \\[1ex]
Present residence since          & 7+ years             & 7+ years             & 7+ years             & 7+ years             & 7+ years             & 7+ years             \\[1ex]
Property                         & no property          & no property          & no property          & insurance            & real estate          & no property          \\[1ex]
Other installment plans          & none                 & none                 & none                 & none                 & none                 & none                 \\[1ex]
Housing                          & rent                 & rent                 & rent                 & rent                 & own                  & own                  \\[1ex]
No. existing credits             & 1                    & 1                    & 1                    & 1                    & 1                    & 1                    \\[1ex]
Job                              & skilled              & skilled              & skilled              & skilled              & skilled              & skilled              \\[1ex]
\textit{No. people being liable} & 0 - 2                & 0 - 2                & 0 - 2                & 0 - 2                & 0 - 2                & 0 - 2                \\[1ex]
Telephone                        & No                   & Yes                  & Yes                  & No                   & No                   & No                   \\[1ex]
\textit{Foreign worker}          & No                   & No                   & No                   & No                   & No                   & No                   \\       
\bottomrule
\end{tabular}
}
\end{table*}

\begin{table*}[!hbt]
\caption{Counterfactual examples from Adult Income dataset.}
\centering
\label{tab:quali_adult}
\vspace{-0.5em}
\resizebox{\linewidth}{!}{
\begin{tabular}{l|r | r | r | r | r | r}
\toprule
\textbf{Adult Income}        & Original input & \multicolumn{5}{c}{Counterfactuals} \\

        & (Low income) & \multicolumn{5}{c}{(High income)} \\

\midrule
Age                     & 18 - 30                & 18 - 30                & 18 - 30                & 18 - 30                & 18 - 30                & 18 - 30                \\[1ex]
Capital gain            & No                     & Yes                    & Yes                    & No                     & No                     & Yes                    \\[1ex]
Capital loss            & No                     & Yes                    & Yes                    & Yes                    & Yes                    & Yes                    \\[1ex]
Hours per week          & 20 - 40                & 20 - 40                & 20 - 40                & 20 - 40                & 20 - 40                & 20 - 40                \\[1ex]
Work class              & Self-employed          & Self-employed          & Self-employed          & Self-employed          & Self-employed          & Self-employed          \\[1ex]
Highest Edu. Level      & 12 & 15 & 14 & 12 & 14 & 12 \\[1ex]

\textit{Marital status} & Married-civ-sps.     & Married-civ-sps.     & Married-civ-sps.     & Married-civ-sps.     & Married-civ-sps.     & Married-civ-sps.     \\
[1ex]

Occupation              & Craft repair           & Other service          & Adm clerical           & Armed forces           & Protective service     & Sales                  \\[1ex]
Relationship            & Not in family          & Wife                   & Other relative         & Wife                & Own child              & Wife                \\[1ex]
\textit{Race}           & Black                  & Black                  & Black                  & Black                  & Black                  & Black                  \\[1ex]
\textit{Sex}            & Female                 & Female                 & Female                 & Female                 & Female                 & Female                 \\[1ex]
\textit{Native country} & Hungary                & Hungary                & Hungary                & Hungary                & Hungary                & Hungary               
\\[1ex]

\bottomrule
\end{tabular}
}
\end{table*}

\begin{table*}[!hbt]
\caption{Counterfactual examples from Graduate Admission dataset.}
\centering
\label{tab:quali_admission}
\vspace{-0.5em}
\resizebox{\linewidth}{!}{
\begin{tabular}{l|r | r | r | r | r | r}
\toprule
\textbf{Graduate Admission}        & Original input & \multicolumn{5}{c}{Counterfactuals} \\

        & (Low chance) & \multicolumn{5}{c}{(High chance)} \\

\midrule
GRE Score                             & 290 - 312 & 290 - 312 & 312 - 323 & 290 - 312 & 290 - 312 & 290 - 312 \\ [1ex]
TOEFL Score                           & 92 - 104  & 104 - 110 & 104 - 110 & 110 - 120 & 110 - 120 & 92 - 104  \\ [1ex]
Undergraduate GPA                     & 7 - 8     & 8 - 9     & 9 - 10    & 9 - 10    & 8 - 9     & 9 - 10    \\ [1ex]
\textit{University Rating} & 2 / 5     & 2 / 5     & 2 / 5     & 2 / 5     & 2 / 5     & 2 / 5     \\ [1ex]
Statement of Purpose                  & 2 / 5     & 2 / 5     & 5 / 5     & 2 / 5     & 2 / 5     & 2 / 5     \\ [1ex]
Letter of Recommendation              & 2.5 / 5   & 4 / 5     & 2.5 / 5   & 2.5 / 5   & 2.5 / 5   & 2 / 5     \\ [1ex]
Research Experience                   & No        & No        & No        & Yes       & No        & Yes      
\\

\bottomrule
\end{tabular}
}
\end{table*}

\begin{table*}[!hbt]
\caption{Counterfactual examples from Student Performance dataset.}
\centering
\label{tab:quali_student}
\vspace{-0.5em}
\resizebox{\linewidth}{!}{
\begin{tabular}{l|r | r | r | r | r | r}
\toprule
\textbf{Student Performance}        & Original input (Fail) & \multicolumn{5}{c}{Counterfactuals (Pass)} \\
\midrule
Age                                 & 16 - 17     & 16 - 17     & 16 - 17     & 16 - 17     & 16 - 17      & 16 - 17     \\ [1ex]
School absences                     & 0 - 4       & 0 - 4       & 0 - 4       & 4 - 32      & 0 - 4        & 4 - 32      \\ [1ex]
\textit{First period grade}         & 10 - 13     & 10 - 13     & 10 - 13     & 10 - 13     & 10 - 13      & 10 - 13     \\[1ex]
Second period grade                 & 0 - 10      & 13 - 19     & 13 - 19     & 13 - 19     & 13 - 19      & 13 - 19     \\[1ex]
\textit{Mother's education}         & Primary     & Primary     & Primary     & Primary     & Primary      & Primary     \\[1ex]
\textit{Father's education}         & None        & None        & None        & None        & None         & None        \\[1ex]
Weekly study time                   & 2 - 5 hours & 2 - 5 hours & 2 - 5 hours & 2 - 5 hours & 5 - 10 hours & 4 - 10 hours \\[1ex]
\textit{Family educational support} & No          & No          & No          & No          & No           & No          \\[1ex]
Wanting higher education            & Yes         & Yes         & No          & No          & Yes          & Yes         \\[1ex]
Internet access at home             & Yes         & No          & Yes         & Yes         & No           & No         \\[1ex]
In a romantic relationship          & Yes         & No          & No          & No          & Yes          & Yes         \\[1ex]
Free time                           & Moderate    & High        & Moderate    & Moderate    & Moderate     & High         \\[1ex]
Going out frequency                 & Very often  & Very often  & Very often  & Rarely      & Very often   & Very Often       \\[1ex]
Health status                       & Good        & Good        & Good        & Very good   & Good         & Good       
   \\[1ex]
\bottomrule
\end{tabular}
}
\end{table*}
\raggedbottom

\end{document}